\pdfoutput=1

\documentclass[sigconf]{acmart}
\AtBeginDocument{%
  }
\usepackage{tcolorbox}
\setcopyright{acmlicensed}
\copyrightyear{2018}
\acmYear{2018}
\acmDOI{XXXXXXX.XXXXXXX}
\acmConference[Conference acronym 'XX]{Make sure to enter the correct
  conference title from your rights confirmation email}{June 03--05,
  2018}{Woodstock, NY}
\acmISBN{978-1-4503-XXXX-X/2018/06}

\usepackage{xspace}
\newcommand{\modelname}{SemRaD\xspace}
\usepackage{pifont}
\usepackage{multirow}
\usepackage{xcolor}

\newcommand{\xmark}{\ding{55}}
\newcommand{\cmark}{\ding{51}}
\newcommand{\suppurl}{https://github.com/lehaoduong123/semrad-supplementary/blob/main/supplementary.pdf}
\newcommand{\suppfootnote}{\footnote{Supplementary Material available at \url{\suppurl}.}}

\begin{document}

\title{Bridging the Information Gap: Semantic Densification and Hindsight Distillation for Cold-Start Prediction}

\author{Hao Duong Le}
\email{lihaoyan25@mails.tsinghua.edu.cn}
\affiliation{%
  \institution{Tsinghua University}
  \city{Beijing}
  \country{China}
}

\author{Yifei Gao}
\email{gao-yf@mail.tsinghua.edu.cn}
\affiliation{%
  \institution{Tsinghua University}
  \city{Beijing}
  \country{China}
}

\author{Huan Li}
\email{huan-li22@mails.tsinghua.edu.cn}
\affiliation{%
  \institution{Tsinghua University}
  \city{Beijing}
  \country{China}
}

\author{Lun Jiang}
\email{jianglun02@meituan.com}
\affiliation{%
  \institution{Meituan}
  \city{Beijing}
  \country{China}
}

\author{Chen Bai}
\email{baichen06@meituan.com}
\affiliation{%
  \institution{Meituan}
  \city{Beijing}
  \country{China}
}

\author{Ke Xing}
\email{xingke@meituan.com}
\affiliation{%
  \institution{Meituan}
  \city{Beijing}
  \country{China}
}

\author{Chen Zhang}
\authornote{Corresponding author.}
\email{zhangchen01@tsinghua.edu.cn}
\affiliation{%
  \institution{Tsinghua University}
  \city{Beijing}
  \country{China}
}

\renewcommand{\shortauthors}{Le et al.}

\begin{abstract}
New-user cold-start is a critical bottleneck for e-commerce platforms: predicting user lifetime value (LTV) and conversion rate (CVR) for users with sparse interaction history. Two prior directions---LLM-based semantic augmentation and learning using privileged information (LUPI)---each face a key limitation. First, LLM augmentation produces \emph{unstructured} rationales that are noisy and hard to operationalize in production. Second, naive \emph{student-teacher} distillation can be brittle due to an information gap between the privileged teacher and the sparse student; moreover, this gap is heterogeneous across users. We propose \modelname{}, a \emph{Semantic Reasoning-aware Distillation} framework addressing both limitations. First, a \emph{Structured Semantic Reasoning Pipeline} replaces free-form rationales with a structured schema built via a \emph{discover--curate--audit} workflow, producing per user a \emph{Densified Semantic Profile} (consumed by the deployed student via a \emph{Semantic-Gated Encoder} that focuses on the most informative dimensions) and a \emph{Hindsight Distillation Target} reconciled from pre- and post-conversion reasoning (used only at training). Second, to bridge this gap and handle its heterogeneity, a \emph{Hindsight-Aware Distillation Network} transfers privileged knowledge via the hindsight target, with \emph{Distillation Experts} improving transfer under per-user variability. On a large-scale industrial dataset, \modelname{} lifts +1.9\% LTV (Gini) and +1.0\% CVR (AUROC) over a production-grade base; a four-week online A/B at Keeta confirms +1.0\% LTV / +0.43\% CVR. \modelname{} also matches the production system's LTV using only 9\% of the training data while improving CVR by 0.8\%.
\end{abstract}


\received{20 February 2007}
\received[revised]{12 March 2009}
\received[accepted]{5 June 2009}

\maketitle

\section{Introduction}

\begin{figure}[t]
    \centering
    \includegraphics[width=\linewidth]{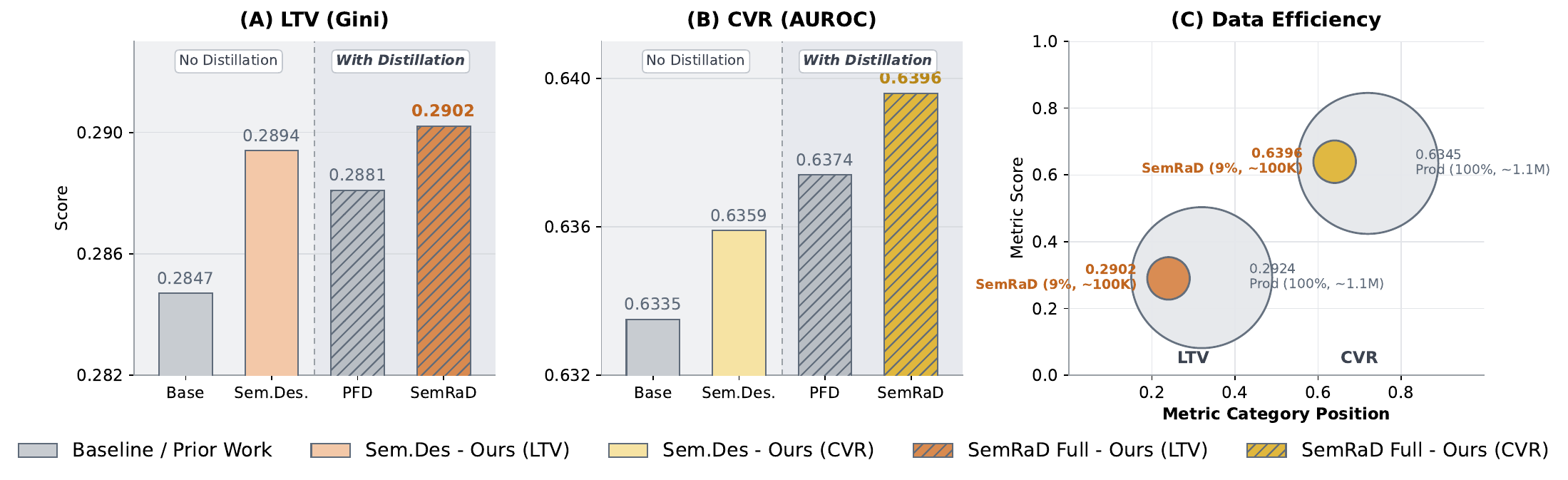}
    \caption{\textbf{Performance and data efficiency of \modelname{}.} (A--B) component ablations; (C) matches production LTV using 9\% of the data (+0.8\% CVR).}
    \Description{Comparison plots showing performance improvements of SemRaD over baselines for LTV and CVR, and a data-efficiency comparison against a production model trained with more data.}
    \label{fig:intro_show_case}
    \vspace{-22px}
\end{figure}

\begin{figure*}[t]
    \centering
    \includegraphics[width=0.78\textwidth]{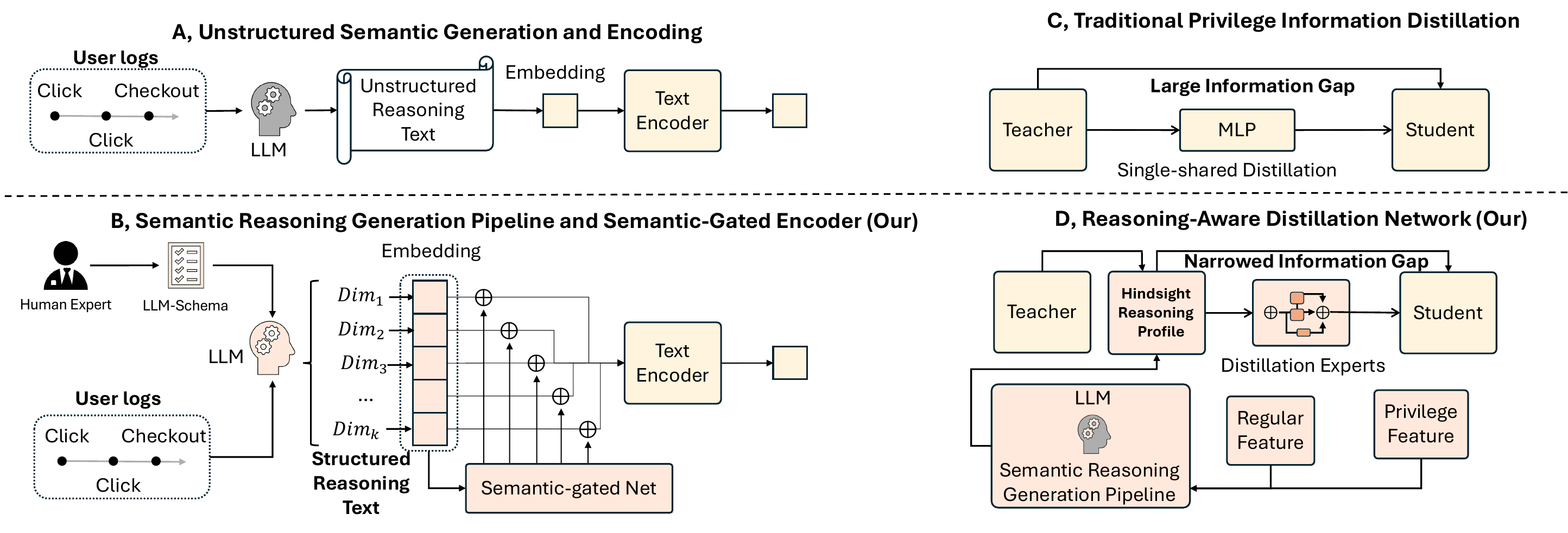}
    \caption{\textbf{Semantic encoding and distillation paradigms.} (A,~C) prior unstructured encoding and large-gap distillation vs.\ (B,~D) our structured schema and hindsight-bridged distillation.}
    \label{fig:method_comparison}
    \vspace{-15px}
\end{figure*}

New-user cold-start is a critical challenge for e-commerce user acquisition (UA) and lifecycle management. Upon signup, platforms must make early lifetime value (LTV) and conversion rate (CVR) predictions for a large number of users with sparse interaction history. For ad and recommendation systems, LTV and CVR are central signals for UA and lifecycle management; extreme sparsity therefore induces high-variance estimates and unreliable ranking decisions~\cite{schein2002coldstart,ma2018esmm}. Standard ID-based sequence models (e.g., DIN~\cite{zhou2018din}, SASRec~\cite{kang2018sasrec}) depend on inductive biases learned from rich histories; when user traces contain only a few events, their predictions often degenerate toward population-level averages, leading to inefficient marketing spend and missed retention opportunities.

Two prior directions have attempted to address this sparsity, each facing a key limitation: (i) \textbf{LLM-based semantic augmentation}, which injects external world knowledge to densify sparse histories~\cite{geng2022p5,wu2023llmrec}, and (ii) \textbf{learning using privileged information (LUPI)}, where a teacher can access post-conversion signals during training~\cite{vapnik2009lupi,lopezpaz2016unifying}.

\emph{First}, LLM-based augmentation can infer high-level user intent (e.g., style preference, functional constraints, brand affinity, browsing depth) from minimal context and densify sparse cold-start traces~\cite{geng2022p5,wu2023llmrec,song2024smartspaces}, with reasoning-centric approaches further eliciting natural-language rationales~\cite{yue2025cot4rec,tsai2024llmreason,kim2025exp3rt}. However, in production pipelines, \emph{unstructured} rationales are difficult to validate, monitor, and operationalize: small prompt rephrasings or market shifts can yield inconsistent explanations even for the same user log~\cite{sclar2024quantifying,ji2023survey}, complicating downstream encoding and ranking. \emph{Second}, LUPI methods distill privileged features (available only during training) from a stronger teacher into a student that uses only online-available signals~\cite{xu2020pfd,yuan2025hapfd}. However, the information gap between teacher and student is often too large for naive transfer to be robust, leading to brittle distillation that fails to inherit the teacher's advantage precisely in the sparse regime that motivates the approach; moreover, this gap is \emph{heterogeneous} across users---the size and pattern of missing signal varies from user to user (Section~\ref{sec:teacher_ablation})---and a single shared distillation pathway averages over these regimes, underfitting exactly the users for whom privileged information is most informative.

To address both limitations together, we propose \modelname{}, a \emph{Semantic Reasoning-aware Distillation} framework (Figure~\ref{fig:method_comparison}). \modelname{} has two complementary components, each addressing one limitation. (i) A \textbf{Structured Semantic Reasoning Pipeline} replaces free-form LLM rationales with structured outputs from a schema built via a \emph{discover--curate--audit} workflow: per user, it produces a \emph{Densified Semantic Profile} that ships with the deployed student, and a \emph{Hindsight Distillation Target} reconciled from pre- and post-conversion reasoning, used only at training. (ii) A \textbf{Hindsight-Aware Distillation Network} then transfers privileged knowledge from teacher to student via the hindsight target, with \emph{Distillation Experts} handling per-user variability. Beyond the cold-start LTV/CVR setting reported here, the framework is being adopted for additional production tasks at Keeta---including CTR prediction, user simulation, and silent-user re-engagement---which share the same structure: a deployable model predicting from sparse signal while richer outcomes are available only at training.

Our contributions are threefold:
\vspace{-1px}

\begin{itemize}
    \item \textbf{Structured Semantic Reasoning Pipeline.} We introduce an LLM-driven pipeline that replaces unstructured rationales with structured outputs from a \emph{discover--curate--audit} schema-design workflow, producing per user a \emph{Densified Semantic Profile} (consumed by the deployed student via a Semantic-Gated Encoder that focuses on the most informative dimensions) and a \emph{Hindsight Distillation Target} reconciled from pre- and post-conversion reasoning.
    \item \textbf{Hindsight-Aware Distillation Network.} We propose a distillation network that bridges the teacher--student information gap via the hindsight target and handles per-user variability via \emph{Distillation Experts}.
    \item \textbf{Generalization \& Production Deployment at Keeta.} We show \modelname{} generalizes across tasks (LTV regression and CVR classification), student backbones, and profiler LLMs, then validate it in production: a four-week online A/B confirms +1.0\% LTV and +0.43\% CVR, and \modelname{} matches the production system's LTV using only 9\% of the training data. The framework is further being adopted for other Keeta production tasks (e.g., CTR, silent-user re-engagement).
\end{itemize}

\section{Related work}
\label{sec:related_work}
\noindent\textbf{\textit{Cold-Start Prediction.}} Standard CVR/LTV backbones~\cite{guo2017deepfm,wang2017dcn,zhou2018din,zhou2019dien,ma2018esmm} excel in warm-start regimes but collapse toward population priors under extreme sparsity. Classical remedies (input dropout~\cite{volkovs2017dropoutnet}, self-supervision) treat the symptom but neither densify sparse inputs with external knowledge nor exploit privileged training-time outcomes---the two levers we target.

\noindent\textbf{\textit{LLM-Based Semantic Augmentation.}} LLMs have been used for recommendation via prompting~\cite{geng2022p5,gao2023chatrec}, fine-tuning~\cite{bao2023tallrec}, and reasoning-centric methods~\cite{yue2025cot4rec,tsai2024llmreason,kim2025exp3rt,wu2023llmrec}, but typically yield \emph{free-form} rationales that are hard to validate or operationalize. We instead generate \emph{structured}, schema-guided reasoning offline and encode it with a frozen embedder for serving efficiency.

\noindent\textbf{\textit{Privileged Information and Distillation.}} LUPI~\cite{vapnik2009lupi,lopezpaz2016unifying}, realized via distillation~\cite{hinton2015distill,romero2015fitnets} and privileged-feature distillation (PFD~\cite{xu2020pfd}, HAPFD~\cite{yuan2025hapfd}), transfers privileged signals through a single shared pathway that is brittle under a heterogeneous gap. We instead add a reconciled hindsight target $\mathbf{Z}_{fused}$ and per-user expert routing~\cite{kang2020de}. Extended discussion is in the \href{\suppurl}{Supplementary Material} (\S\,S5).

\section{Problem Formulation}
\label{sec:problem_formulation}

We consider new-user cold-start prediction anchored at each user's first-conversion event $t_0(u)$ (i.e., the timestamp of their initial purchase).
Let $\mathcal{E}_u(t)$ denote the stream of user interactions (e.g., views, clicks, add-to-cart) up to time $t$.

\paragraph{Pre- vs. post-conversion logs (training vs. serving).}
At serving time, the model only observes all interaction logs \emph{before} the first conversion:
\begin{equation}
    \mathbf{S}_{\text{pre}}(u) \triangleq \mathcal{E}_u\big((-\infty,\, t_0(u))\big).
\end{equation}
During training, we additionally have access to the post-conversion log within a fixed 30-day window after the first conversion as privileged information:
\begin{equation}
    \mathbf{S}_{\text{post}}(u) \triangleq \mathcal{E}_u\big([t_0(u),\, t_0(u)+30\text{d}]\big).
\end{equation}
We use $\mathbf{S}_{\text{pre}}(u)$ and $\mathbf{S}_{\text{post}}(u)$ as the model-ready feature sequences throughout.

\paragraph{Prediction targets.}
We predict two downstream business metrics defined on a fixed 30-day window after the first conversion:
\begin{equation}
\begin{split}
    y^{\text{LTV}}(u) &\triangleq \text{GMV of Post Orders}\big(u;\, [t_0(u),\, t_0(u)+30\text{d}]\big), \\
    y^{\text{CVR}}(u) &\triangleq \mathbb{1}\big[\text{Post Orders Exist in } [t_0(u),\, t_0(u)+30\text{d}]\big].
\end{split}
\end{equation}
Here, $y^{\text{CVR}}$ indicates whether the user makes \emph{any post-conversion purchase} within 30 days after their first conversion, and $y^{\text{LTV}}$ measures the total Gross Merchandise Value (GMV) of all post-conversion orders within the 30-day window (set to 0 if no post-conversion orders occur).
For each task $\tau \in \{\text{LTV}, \text{CVR}\}$, the goal is to learn a separate student model that satisfies the serving constraint
\begin{equation}
    \hat{y}^{\tau}(u)=f_{\theta_\tau}\big(\mathbf{S}_{\text{pre}}(u)\big), \quad \tau \in \{\text{LTV}, \text{CVR}\},
\end{equation}
while training may utilize $\mathbf{S}_{\text{post}}(u)$ through a privileged teacher. The two tasks are trained separately on the same augmented data.

\section{Structured Semantic Reasoning Pipeline}
\label{sec:pipeline}
\begin{figure*}[t]
    \centering
    \includegraphics[width=0.82\textwidth]{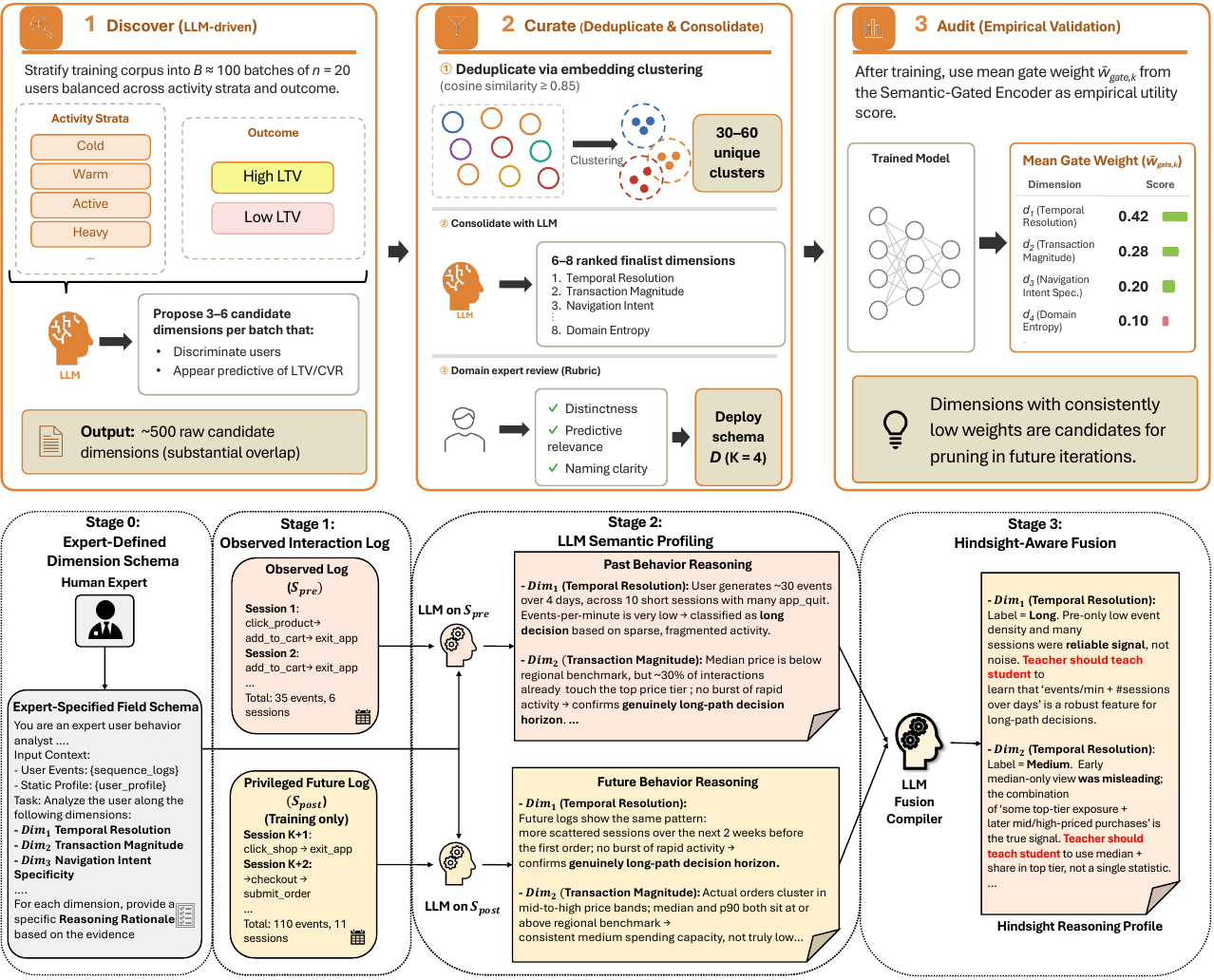}
    \caption{\textbf{Structured Semantic Reasoning Pipeline} (Stages 0--3): schema construction via \emph{discover--curate--audit}, log collection ($\mathbf{S}_{pre}$/$\mathbf{S}_{post}$), LLM profiling ($\mathbf{Z}_{pre}$/$\mathbf{Z}_{post}$), and hindsight fusion into $\mathbf{Z}_{fused}$.}
    \vspace{-10px}
    \label{fig:data_pipeline}
\end{figure*}

We address the cold-start information deficit by employing Large Language Models (LLMs) to project sparse user histories into a dense, interpretable semantic space (Figure~\ref{fig:data_pipeline}). Our pipeline involves three LLM calls under a structured schema: (i) profiling $\mathbf{S}_{pre}$ to generate the \emph{Densified Semantic Profile} $\mathbf{Z}_{pre}$ (available at inference), (ii) profiling $\mathbf{S}_{post}$ to generate the privileged future profile $\mathbf{Z}_{post}$ (training only), and (iii) fusing $\mathbf{Z}_{pre}$ and $\mathbf{Z}_{post}$ to produce the \emph{Hindsight Distillation Target} $\mathbf{Z}_{fused}$ (training only).

\subsection{Stage 0: Schema Construction via Discover--Curate--Audit}
\label{sec:stage0_dimension_discovery}

As discussed in the introduction, unstructured LLM-generated rationales are difficult to validate, monitor, and operationalize in production. We therefore introduce a task-specific behavioral dimension schema $\mathcal{D}=\{D_1,\ldots,D_K\}$ that structures the LLM's reasoning along predefined, interpretable axes (Figure~\ref{fig:data_pipeline}, Stage 0), where each dimension $D_k$ captures a distinct aspect of user behavior identified as predictive of conversion (e.g., \textit{Temporal Resolution}, \textit{Transaction Magnitude}, \textit{Navigation Intent Specificity}, \textit{Domain Entropy}). Constraining the LLM to these axes makes the reasoning (i) \emph{interpretable} and verifiable against business intuition, (ii) \emph{consistent} across users and time, and (iii) \emph{task-relevant} rather than arbitrary narrative.

\noindent\textbf{\textit{Schema construction.}} Hand-engineering this schema is brittle, while ad-hoc LLM discovery risks irrelevant or unstable dimensions. We therefore adopt a three-step \emph{discover--curate--audit} workflow that combines LLM-driven discovery, light human oversight, and a built-in empirical validation step.

\noindent\textbf{\textit{Discover.}} We stratify the training corpus into $B\approx100$ batches of $n{=}20$ users each, balanced across activity strata (cold/warm/active/heavy) and outcome (high/low LTV). For each batch, an LLM proposes 3--6 candidate dimensions that discriminate the users and appear predictive of LTV/CVR, yielding $\sim$500 raw candidates with substantial overlap.

\noindent\textbf{\textit{Curate.}} Candidates are deduplicated via embedding clustering (cosine $\geq0.85$) into 30--60 unique clusters, then consolidated by a second LLM call into 6--8 ranked finalists. Domain experts review these against a rubric checking distinctness, predictive relevance, and naming clarity, requiring domain familiarity rather than ML expertise; the deployed schema $\mathcal{D}$ contains $K{=}4$ dimensions.

\noindent\textbf{\textit{Audit.}} After training, the mean gate weight $\bar{w}_{\mathrm{gate},k}$ from the Semantic-Gated Encoder serves as an empirical utility score for each dimension; consistently down-weighted dimensions are candidates for pruning. Section~\ref{sec:gating_analysis} reports per-dimension utilization for the deployed schema. Discovery and consolidation prompts, and dimension examples, are in the Supplementary Material (\S\,S4)\suppfootnote.

\subsection{Stage 1: Observed Interaction Log}
\label{sec:stage1_observed_log}
We collect two behavioral logs per user (Figure~\ref{fig:data_pipeline}, Stage 1): the observed pre-conversion log $\mathbf{S}_{pre}$ (available at serving time, before first-conversion $t_0$) and the privileged future log $\mathbf{S}_{post}$ (post-$t_0$, training only).

\subsection{Stage 2: LLM Semantic Profiling (Pre vs. Post)}
\label{sec:student_densification}
As shown in Figure~\ref{fig:data_pipeline} (Stage 2), we apply a profiling prompt $\mathcal{P}_{\text{prof}}$ to generate structured reasoning along the dimensions $\mathcal{D}$ defined in Stage 0 (prompt details in the \href{\suppurl}{Supplementary Material}, \S\,S2).

For the student, we profile the pre-conversion log $\mathbf{S}_{pre}$ to obtain the \emph{Densified Semantic Profile} $\mathbf{Z}_{pre} = \{\mathbf{t}_k^{pre}\}_{k=1}^K$, where each $\mathbf{t}_k^{pre}$ is a natural language rationale explaining the user's behavior along dimension $D_k$. This profile is the only LLM-derived signal consumed by the deployed student at serving time. For the teacher (training only), we apply the same profiling process to the privileged post-conversion log $\mathbf{S}_{post}$ to obtain the privileged future profile $\mathbf{Z}_{post} = \{\mathbf{t}_k^{post}\}_{k=1}^K$.

\subsection{Stage 3: Hindsight-Aware Fusion}
\label{sec:teacher_hindsight}

Prior privileged-feature distillation~\cite{xu2020pfd,yuan2025hapfd} simply provides the teacher with both $\mathbf{Z}_{pre}$ and $\mathbf{Z}_{post}$. However, when these contradict, the teacher receives inconsistent supervision that is difficult for the student to learn from. We instead apply a fusion prompt $\mathcal{P}_{\text{fusion}}$ (Figure~\ref{fig:data_pipeline}, Stage 3) that, for each dimension $D_k$, compares the pre-conversion hypothesis with the post-conversion outcome, resolves contradictions by identifying which pre-conversion signals were truly predictive, and outputs a per-dimension hindsight target $\mathbf{t}_k^{teacher}$. The complete \emph{Hindsight Distillation Target} is $\mathbf{Z}_{fused} = \{\mathbf{t}_k^{teacher}\}_{k=1}^K$. During training, the teacher consumes only $\mathbf{Z}_{fused}$. Because $\mathbf{Z}_{fused}$ already carries the salient pre- and post-conversion signals together with the explicit reconciliation between them, it provides the teacher with self-consistent supervision---rather than the raw, potentially contradictory $\mathbf{Z}_{pre}$ and $\mathbf{Z}_{post}$---that the student (which sees only $\mathbf{Z}_{pre}$) can effectively learn from. We validate this design via prediction disagreement analysis in Section~\ref{sec:teacher_ablation}.

\section{Hindsight-Aware Distillation Network}
\begin{figure*}[]
    \centering
    \includegraphics[width=0.82\textwidth]{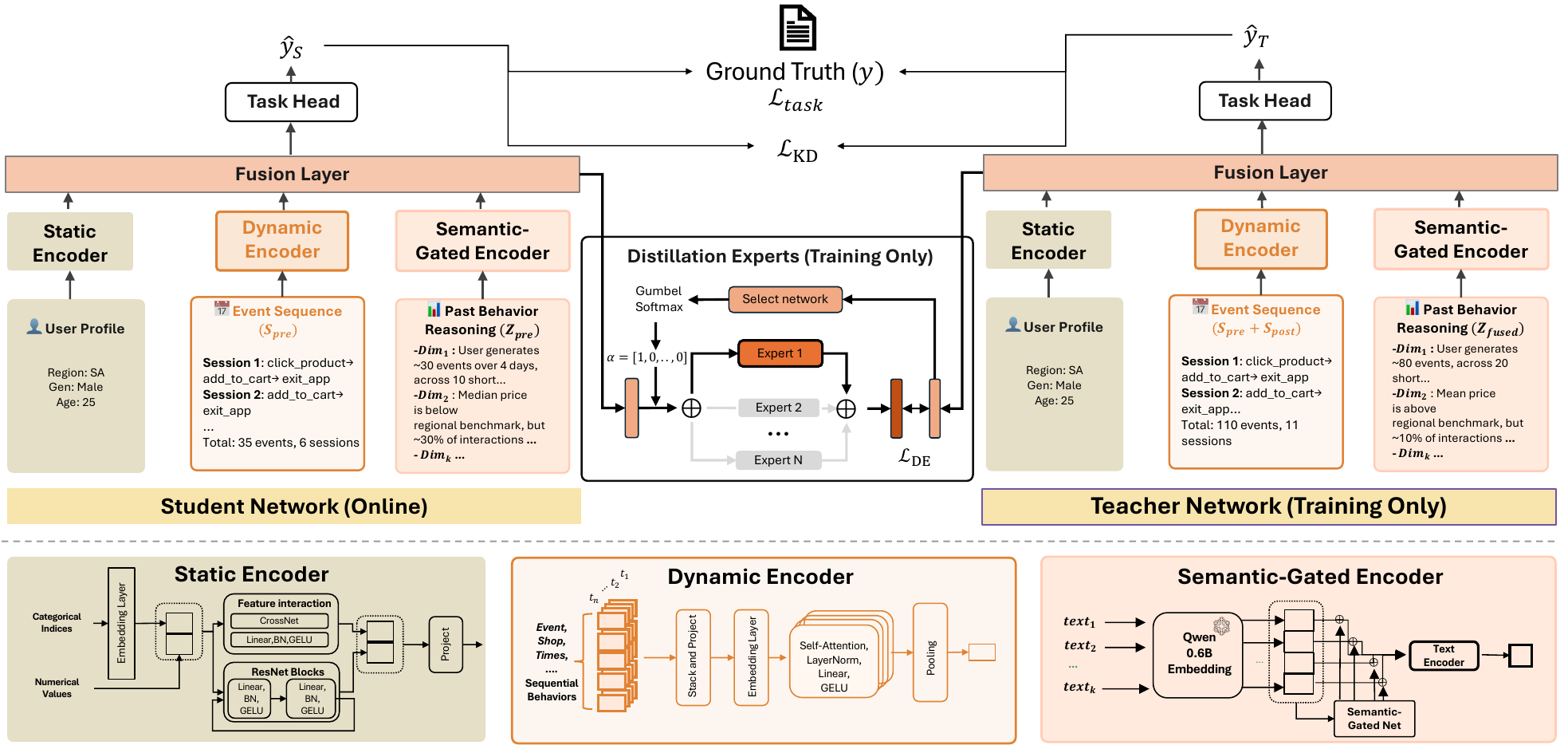}
    \caption{\textbf{Hindsight-Aware Distillation Network.} Shared student/teacher encoders (Static, Dynamic, Semantic-Gated); student sees ($\mathbf{S}_{pre}$, $\mathbf{Z}_{pre}$), teacher sees ($\mathbf{S}_{pre}\oplus\mathbf{S}_{post}$, $\mathbf{Z}_{fused}$). Distillation Experts ($\mathcal{L}_{DE}$) and logit KD ($\mathcal{L}_{kd}$) transfer privileged knowledge.}
    \vspace{-10px}
    \label{fig:architecture}
\end{figure*}

The proposed \modelname{} framework follows a \textbf{Simultaneous Distillation} paradigm \citep{xu2020pfd}. We instantiate two identical networks with shared architecture: a Student $\mathcal{S}$ and a Teacher $\mathcal{T}$. They differ only in their inputs: $\mathcal{S}$ observes $\mathbf{S}_{pre}$, while $\mathcal{T}$ observes both the regular and privileged data $\mathbf{S}_{pre} \oplus \mathbf{S}_{post}$. Both models are trained jointly. Crucially, our framework is \textbf{Model-Agnostic}. It is designed as a plug-and-play enhancement that can be integrated with any standard industrial backbone (DIN or Transformer). 

\subsection{Multi-View Feature Encoding}
\label{sec:encoders}

The Student and Teacher share encoder architectures across three views and differ only in inputs. The \textbf{Static view} embeds categorical and numerical user-profile features (shop ID, brand ID, tenure, session counts) via embedding lookups + standardized numerics, projected by an MLP to $\mathbf{h}_{stat} \in \mathbb{R}^{d_{stat}}$; both networks share this view since no privileged signal exists at the profile level. The \textbf{Dynamic view} encodes the event sequence (event type, shop type, shop AOI; AOV, price, time gap) with a Transformer followed by attention-pooling:
\begin{equation}
    \mathbf{h}_{dyn} = \text{AttnPool}(\text{Transformer}(\{e_1, \ldots, e_T\})) \in \mathbb{R}^{d_{dyn}}
\end{equation}
The Student encodes $\mathbf{S}_{pre}$, while the Teacher encodes $\mathbf{S}_{pre} \oplus \mathbf{S}_{post}$~\cite{xu2020pfd}. The \textbf{Semantic Reasoning view} consumes the rationales from Section~\ref{sec:pipeline}: the Student receives $\mathbf{Z}_{pre}$ and the Teacher receives the reconciled $\mathbf{Z}_{fused}$, both encoded by the Semantic-Gated Encoder below to produce $\mathbf{h}_{sem}^{S}$ and $\mathbf{h}_{sem}^{T}$. Feature details are in the \href{\suppurl}{Supplementary Material} (\S\,S3).

\subsection{Semantic-Gated Encoder}
\label{sec:semantic_gating}

User behavior is heterogeneous: different semantic dimensions matter for different users (e.g., \textit{Temporal Resolution} dominates for rapid converters, \textit{Domain Entropy} for diverse browsers). Naively concatenating all $K$ semantic embeddings introduces interference from irrelevant dimensions. We instead encode each rationale $\mathbf{t}_k$ with a frozen text embedder (Qwen-0.6B-Emb) and compute per-user importance weights:
\begin{equation}
    \mathbf{e}_{k} = \text{TextEmbed}(\mathbf{t}_k), \quad \mathbf{w}_{gate} = \text{Softmax}(\text{MLP}_{gate}([\mathbf{e}_{1}, \ldots, \mathbf{e}_{K}]))
\end{equation}
The final semantic representation is the gate-weighted aggregation:
\begin{equation}
    \mathbf{h}_{sem} = \text{MLP}_{sem}([w_{gate,1}\mathbf{e}_{1}, \ldots, w_{gate,K}\mathbf{e}_{K}]) \in \mathbb{R}^{d_{sem}}
\end{equation}
The gate weights $\mathbf{w}_{gate}$ also provide interpretability and serve as a post-hoc empirical utility score for the expert-defined dimensions (Section~\ref{sec:stage0_dimension_discovery}); per-dimension utilization is reported in Section~\ref{sec:gating_analysis}.

\subsection{Distillation Experts}
\label{sec:distillation_experts}

Prior latent-alignment approaches (e.g., HAPFD~\cite{yuan2025hapfd}) transfer privileged knowledge through a \emph{single shared projection pathway}. This assumes that one student-to-teacher mapping is sufficient for all users, even though the teacher--student information gap can vary across users: for some users, post-conversion evidence may clarify pre-conversion ambiguity, while for others it may introduce signals that are difficult to infer from the student's inputs. To avoid forcing these heterogeneous transfer patterns into one pathway, we adopt \emph{Distillation Experts}~\cite{kang2020de}: $M$ expert MLPs whose outputs are mixed by a Gumbel-softmax router conditioned on the teacher's pooled representation, producing an expert-routed reconstruction of the teacher target:
\begin{equation}
    \hat{\mathbf{h}}_T = \sum_{m=1}^M \alpha_m(\text{sg}(\mathbf{h}_{T,dyn})) \cdot \text{Expert}_m(\mathbf{h}_{S,dyn}), \quad \mathcal{L}_{DE} = \| \hat{\mathbf{h}}_T - \text{sg}(\mathbf{h}_{T,dyn}) \|^2_2
\end{equation}
where $\alpha_m(\cdot) = \text{GumbelSoftmax}(\text{MLP}_{router}(\cdot))_m$ are the per-user routing weights. Following the original formulation~\cite{kang2020de}, the router conditions on the (detached) teacher representation, allowing privileged information to choose among multiple reconstruction pathways; the experts themselves operate on the \emph{student} representation $\mathbf{h}_{S,dyn}$, so the reconstruction must remain achievable from $\mathbf{S}_{pre}$. The experts and router are used only during training; the deployed student requires only its standard forward pass, so this added capacity incurs no serving cost.

\subsection{Training Objective}
\label{sec:optimization}

\modelname{} is trained \emph{per task}: we optimize a single task loss at a time---Huber regression for LTV or Binary Cross-Entropy for CVR---training the two tasks as separate models that share the same augmented data ($\mathbf{Z}_{pre}$, $\mathbf{Z}_{fused}$). For a given task, the Student and Teacher are trained jointly with a weighted sum of three terms:
\begin{equation}
    \mathcal{L}_{total} = \underbrace{\mathcal{L}_{task}(\mathcal{S}) + \mathcal{L}_{task}(\mathcal{T})}_{\text{Supervised}} + \underbrace{\lambda_{kd}\mathcal{L}_{kd}}_{\text{Soft-Target KD}} + \underbrace{\lambda_{DE}\mathcal{L}_{DE}}_{\text{Experts}}
\end{equation}
where $\mathcal{L}_{task}$ is the task-specific supervised loss (Huber for LTV, Binary Cross-Entropy for CVR); $\mathcal{L}_{kd}$ transfers the teacher's soft targets to the student (MSE for LTV regression, temperature-scaled KL~\cite{hinton2015distill} for CVR classification); and $\mathcal{L}_{DE}$ is the expert-routed reconstruction loss from Section~\ref{sec:distillation_experts}. By minimizing $\mathcal{L}_{total}$, the Teacher learns from the privileged future ($\mathbf{S}_{post}$) and pulls the Student towards this privileged manifold via both soft targets and per-user expert-routed reconstruction.

\section{Experiments}
\label{sec:experiments}
\subsection{Experimental Setup}
\begin{table*}[t]
\centering
\caption{Method comparison and \modelname{} ablations (mean $\pm$ std). Top: augmentation/distillation strategies; bottom: loss-component ablations. $\mathbf{Z}_{pre}$/$\mathbf{Z}_{post}$/$\mathbf{Z}_{fused}$: densified\,/\,privileged\,/\,hindsight profiles.}
\label{tab:prompt_ablation}
\begin{tabular}{@{}lccccccc@{}}
\toprule
\textbf{Model / Variant} & \textbf{Distill} & $\mathbf{Z}_{pre}$ & $\mathbf{Z}_{post}$ & $\mathbf{Z}_{fused}$ & \textbf{CVR (AUROC)} & \textbf{LTV (Gini)} \\
\midrule
(A) Base & \xmark & \xmark & \xmark & \xmark & 0.6335 $\pm$ 0.0056 & 0.2847 $\pm$ 0.0096 \\
(B) Unstructured~\cite{kojima2023largelanguagemodelszeroshot} & \xmark & \cmark & \xmark & \xmark & 0.6308 $\pm$ 0.005 & 0.2826 $\pm$ 0.005 \\ 
(C) Few-shot & \xmark & \cmark & \xmark & \xmark & 0.6321 $\pm$ 0.004 & 0.2841 $\pm$ 0.005 \\ 
(D) Sem.\ Des.\ (Ours) & \xmark & \cmark & \xmark & \xmark & 0.6359 $\pm$ 0.0039 & 0.2894 $\pm$ 0.0084 \\
\midrule
(E) Self-distillation & \cmark & \cmark & \xmark & \xmark & 0.6362 $\pm$ 0.0029 & 0.2889 $\pm$ 0.0056 \\
(F) PFD~\cite{xu2020pfd} & \cmark & \cmark & \cmark & \xmark & 0.6374 $\pm$ 0.0025 & 0.2881 $\pm$ 0.0032 \\
(G) HAPFD~\cite{yuan2025hapfd} & \cmark & \cmark & \cmark & \xmark & 0.6353 $\pm$ 0.0033 & 0.2858 $\pm$ 0.0068 \\
\textbf{\modelname{} (Ours)} & \cmark & \xmark & \xmark & \cmark & \textbf{0.6396 $\pm$ 0.0027} & \textbf{0.2902 $\pm$ 0.0045} \\
\midrule
\multicolumn{7}{@{}l}{\emph{\modelname{} ablations (share \modelname{} input configuration):}} \\
\quad w/o Semantic Gating & --- & --- & --- & --- & 0.6362 $\pm$ 0.0025 & 0.2834 $\pm$ 0.0038 \\ 
\quad Logit KD only (no DE)   & --- & --- & --- & --- & 0.6374 $\pm$ 0.0025 & 0.2881 $\pm$ 0.0032 \\ 
\quad Distillation Experts only (no KD) & --- & --- & --- & --- & 0.6362 $\pm$ 0.0029 & 0.2889 $\pm$ 0.0056 \\ 
\bottomrule
\end{tabular}
\end{table*}

\noindent\textbf{\textit{Dataset.}}
We use a real-world industrial dataset (collected June--August) with a strictly temporal split of 120,000 stratified users (100k/10k/10k train/val/test). Each user is anchored at first-conversion time $t_0$ with observed $\mathbf{S}_{pre}$ and privileged $\mathbf{S}_{post}$ (training only), described by static and dynamic sequence features (\href{\suppurl}{Supplementary Material}, \S\,S3).

\noindent\textbf{\textit{Metrics.}}
We report the Normalized Gini Coefficient for LTV (regression; robust to label scale, emphasizing high-value ranking) and AUROC for CVR (classification).

\noindent\textbf{\textit{Baselines.}}
All methods share the inference-time log $\mathbf{S}_{pre}$ and differ only in training-time augmentation and distillation (configurations in Table~\ref{tab:prompt_ablation}). \emph{Augmentation-only:} (A) Base, (B) Unstructured~\cite{kojima2023largelanguagemodelszeroshot}, (C) Few-shot, (D) Sem.\ Des.\ (Ours). \emph{Distillation:} (E) Self-distillation (teacher sees only $\mathbf{Z}_{pre}$, isolating distillation architecture from privileged signal), (F) PFD~\cite{xu2020pfd}, (G) HAPFD~\cite{yuan2025hapfd} (teacher sees $\mathbf{Z}_{pre} \oplus \mathbf{Z}_{post}$). \modelname{} (Ours) adds the reconciled target $\mathbf{Z}_{fused}$ and Distillation Experts.

\noindent\textbf{\textit{Evaluation Protocol.}}
We organize our evaluation around four questions: \textbf{RQ1}---do the two components each improve cold-start prediction, and are they complementary (Section~\ref{sec:main_results}); \textbf{RQ2}---why does hindsight fusion help, and does its benefit concentrate on cold users; \textbf{RQ3}---does \modelname{} generalize across tasks, backbones, and profiler LLMs; and \textbf{RQ4}---is \modelname{} deployable, i.e.\ data-efficient and effective in production. Extended results and additional ablations are provided in the \href{\suppurl}{Supplementary Material}.

\noindent\textbf{\textit{Implementation.}}
\modelname{} (PyTorch) combines a CrossNet-based Static Encoder, a Transformer Dynamic Encoder (2 layers, 4 heads), and a Semantic-Gated Encoder over Qwen-0.6B embeddings; profiles are generated by GPT-OSS 120B. Full architecture and training hyperparameters are in the \href{\suppurl}{Supplementary Material} (\S\,S1).

\subsection{Main Results}
\label{sec:main_results}

Table~\ref{tab:prompt_ablation} summarizes the main comparison. Structured semantic descriptions improve over the base and unstructured augmentation baselines, while the full \modelname{} model achieves the best results on both tasks: $+1.9\%$ LTV Gini and $+1.0\%$ CVR AUROC over Base. The lower block gives a compact component check: removing semantic gating or using only one distillation loss weakens the full recipe, although the KD-vs.-expert differences are close to seed variance. We therefore use the table as the primary quantitative evidence and focus the remaining main-paper analysis on the two learned mechanisms that make the method interpretable.

\begin{figure}[ht]
    \vspace{-6pt}
    \centering
    \includegraphics[width=0.42\textwidth]{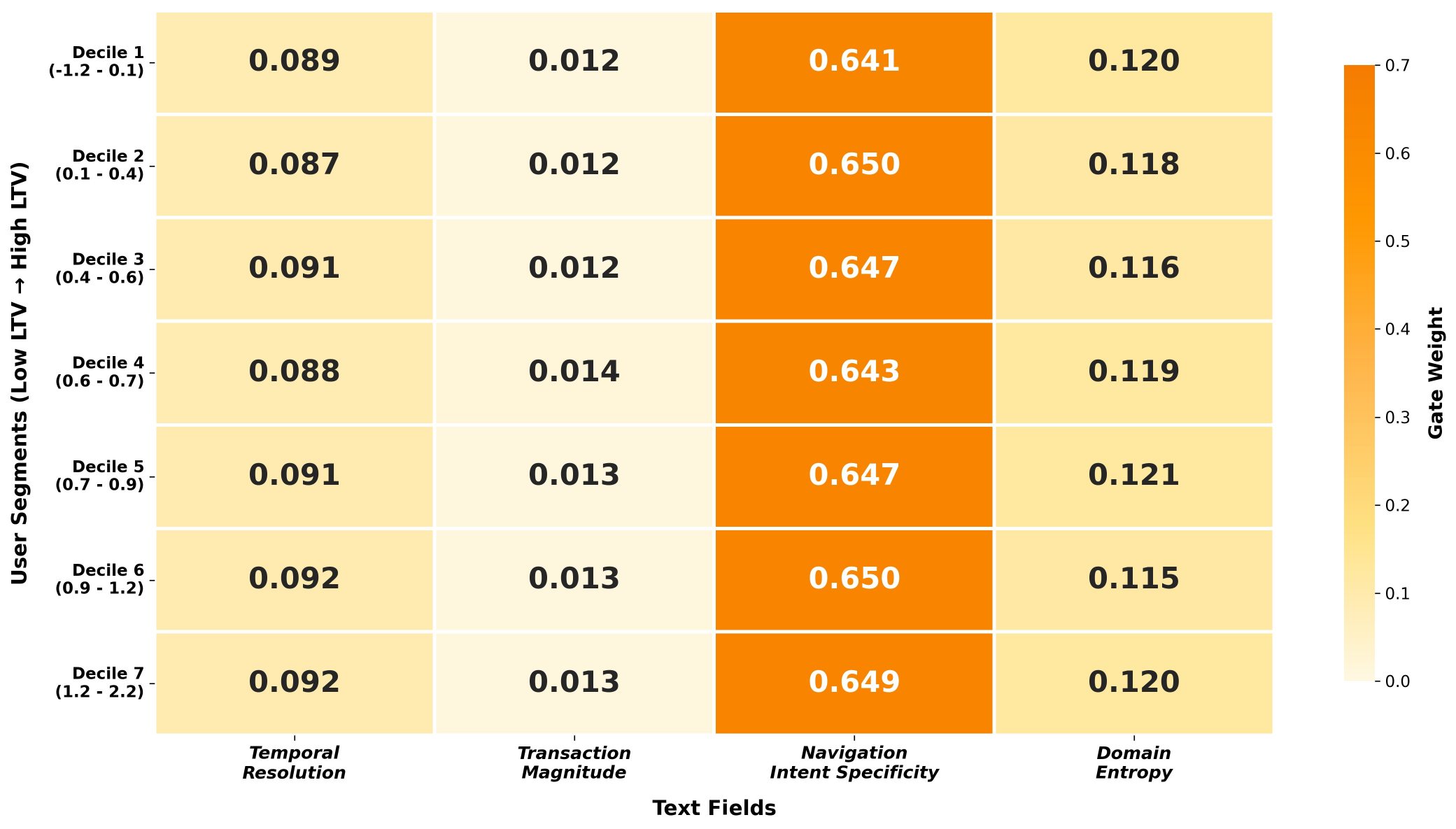}
    \caption{\textbf{Semantic gating weights} per dimension, by LTV decile.}
    \Description{Bar chart of average semantic gate weights across LTV user segments, showing dimension-selective schema utilization.}
    \label{fig:gating_weights}
    \vspace{-10px}
\end{figure}

\noindent\textbf{\textit{Schema Utility via Gate Weights.}}
\label{sec:gating_analysis}
The learned gate weights of the Semantic-Gated Encoder provide a post-hoc empirical utility score for each schema dimension (Section~\ref{sec:stage0_dimension_discovery})---the \emph{Audit} step of the discover--curate--audit workflow. Figure~\ref{fig:gating_weights} shows that the gate is strongly dimension-selective---Navigation Intent Specificity dominates across all user segments while Transaction Magnitude is nearly suppressed---and that this ordering matches expert intuition. No dimension is consistently down-weighted enough to warrant pruning, confirming the schema is both interpretable and well-chosen.

\begin{figure}[ht]
    \vspace{-6pt}
    \centering
    \includegraphics[width=0.42\textwidth]{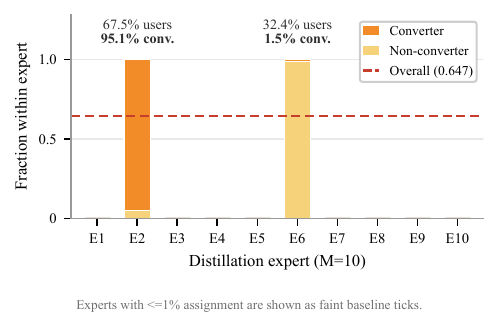}
    \caption{\textbf{Post-hoc routing of Distillation Experts.} Of $M{=}10$ experts, two dominate ($>1\%$ assignment) and capture opposite conversion-outcome regimes.}
    \Description{Stacked vertical bars show within-expert conversion composition for ten configured experts. Two experts dominate, capturing opposite conversion-outcome regimes.}
    \label{fig:expert_routing}
    \vspace{-10px}
\end{figure}

\noindent\textbf{\textit{Post-hoc Expert Routing.}}
Inspecting the learned Distillation Expert router, we find that---without any group supervision---it converges to two dominant experts aligned with conversion outcome rather than with cold/warm or LTV strata (Figure~\ref{fig:expert_routing}): one pathway handles users whose privileged window confirms purchase, the other handles users with little post-conversion signal.

\noindent\textbf{\textit{Information Gap Analysis.}}
\label{sec:teacher_ablation}
Beyond the post-hoc routing behavior, we ask whether hindsight fusion also makes privileged supervision more reachable for the student. To answer RQ2, we quantify the information gap by measuring student-teacher prediction disagreement across user history buckets. Lower disagreement indicates higher \emph{distillability}: when the privileged teacher's predictions are closer to what the student can achieve from sparse pre-conversion inputs, the teacher provides more transferable supervision rather than unattainable targets. We measure disagreement as absolute LTV difference $|y_T-y_S|$ and CVR logit difference $|\text{logit}(p_T)-\text{logit}(p_S)|$.

\begin{figure}[ht]
    \centering
    \includegraphics[width=0.42\textwidth]{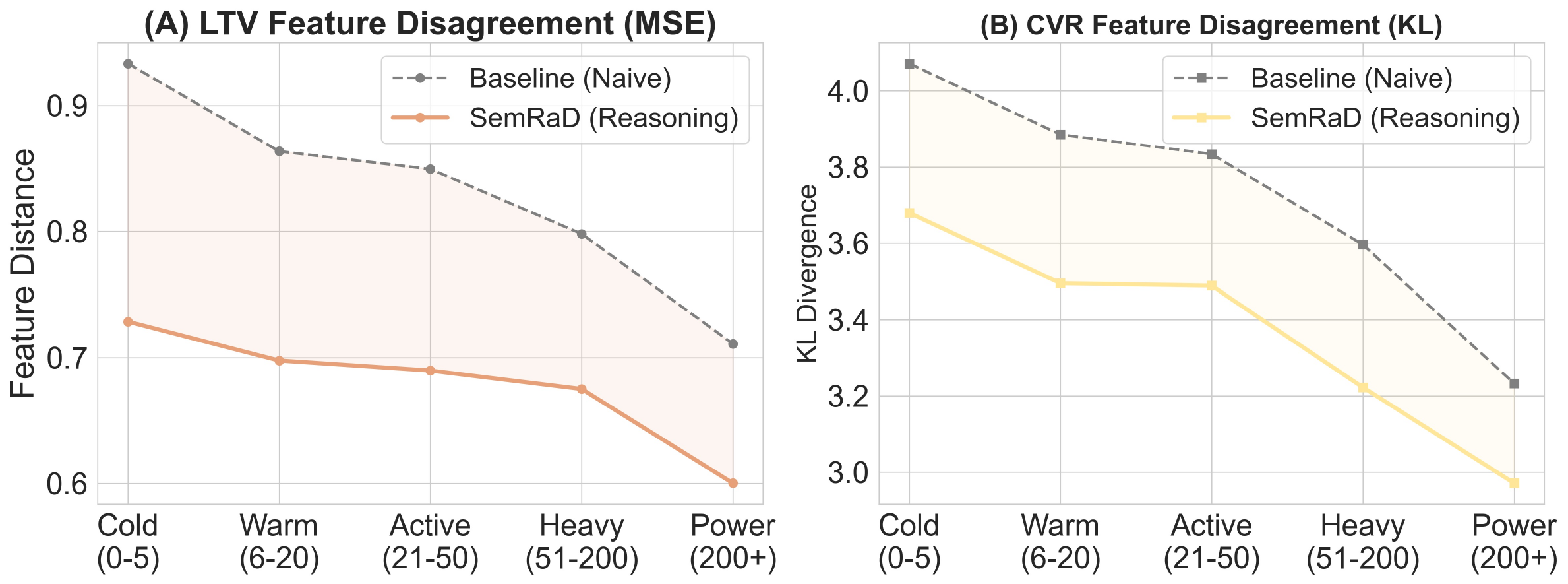}
    \caption{\textbf{Student-Teacher Predictive Disagreement.} \modelname{} reduces disagreement for cold-start users, improving distillability.}
    \Description{Line plots compare student-teacher prediction disagreement across user-history buckets for LTV and CVR. SemRaD shows lower disagreement than naive privileged distillation, especially for cold-start users.}
    \label{fig:gap_analysis}
    \vspace{-10px}
\end{figure}

Figure~\ref{fig:gap_analysis} shows that naive privileged distillation has the largest mismatch for cold-start users ($N \le 5$), with an LTV gap of $0.933$ and a CVR logit gap of $4.07$. \modelname{} reduces these gaps to $0.728$ ($-22\%$) and $3.67$ ($-10\%$), respectively. This supports our hypothesis that hindsight fusion does not simply add more privileged information; it reconciles post-conversion evidence with pre-conversion signals, making the teacher's guidance more reachable for the student, especially in the sparse-history regime.

\subsection{Generalization and Robustness}
\label{sec:robustness}

\noindent\textbf{\textit{RQ3. Does \modelname{} generalize across tasks, backbones, and profiler LLMs?}}
\modelname{} generalizes along two concrete axes (Table~\ref{tab:robustness}). First, replacing the Transformer sequence encoder with Avg-Pooling or DIN preserves the improvement, with the attention-based DIN benefiting most. Second, gains remain under alternative profiler LLMs, indicating that the effect comes from the structured semantic schema and hindsight fusion rather than a single upstream generator.

\begin{table}[h]
    \centering
    \caption{Robustness across student backbones and profiler LLMs (Transformer/GPT-OSS is the main-table setting).}
    \vspace{-8px}
    \label{tab:robustness}
    \begin{tabular}{@{}lcc@{}}
    \toprule
    \textbf{Configuration} & \textbf{LTV (Gini)} & \textbf{CVR (AUROC)} \\
    \midrule
    \multicolumn{3}{@{}l}{\emph{(a) Student backbone (w/o $\rightarrow$ w/ \modelname{})}} \\
    \quad Avg-Pooling & $.2710\!\rightarrow\!\mathbf{.2832}$ & $.6282\!\rightarrow\!\mathbf{.6382}$ \\
    \quad DIN         & $.2699\!\rightarrow\!\mathbf{.2898}$ & $.6268\!\rightarrow\!\mathbf{.6428}$ \\
    \quad Transformer & $.2847\!\rightarrow\!\mathbf{.2902}$ & $.6335\!\rightarrow\!\mathbf{.6396}$ \\
    \midrule
    \multicolumn{3}{@{}l}{\emph{(b) Profiler LLM (Transformer backbone)}} \\
    \quad GPT-OSS 120B (ours)   & $\mathbf{.2902\pm.0045}$ & $\mathbf{.6396\pm.0027}$ \\
    \quad Qwen3.5-9B            & $.2834\pm.0042$ & $.6374\pm.0019$ \\
    \quad Qwen3.5-35B-A3B (MoE) & $.2820\pm.0098$ & $.6360\pm.0052$ \\
    \bottomrule
    \end{tabular}
\end{table}

\subsection{Production Deployment}
\label{sec:deployment}

\noindent\textbf{\textit{RQ4. Is \modelname{} deployable---data-efficient and effective in production?}}
\noindent\textbf{\textit{Data Efficiency.}}
LTV labels are expensive and slow to collect, so we first ask whether structured reasoning can reduce dependence on large-scale supervision. In a small-data setting, \modelname{} reaches production-level LTV quality using only \textbf{9\% of the training data} while improving CVR by 0.8\% (Figure~\ref{fig:intro_show_case}C), suggesting that the semantic profiles provide useful signal for new markets and other cold-start regimes.

\noindent\textbf{\textit{Online A/B Test.}}
We deployed \modelname{} to live traffic against the production baseline. Table~\ref{tab:online_deployment} reports the headline A/B results together with deployment cost and latency. \modelname{} improves both online LTV ranking and CVR discrimination, while serving uses only the cached frozen embedding plus the gated MLP---no LLM, teacher, or fusion call is needed at inference time.

\begin{table}[t]
    \centering
    \caption{Online A/B at Keeta (4 weeks; 5\% traffic; 10{,}000 users / 6{,}791 conversions).}
    \vspace{-6px}
    \label{tab:online_deployment}
    \begin{tabular}{@{}lccc@{}}
    \toprule
     & \textbf{Production} & \textbf{\modelname{}} & \textbf{Rel.\ gain} \\
    \midrule
    \multicolumn{4}{@{}l}{\textit{Outcome quality}} \\
    \quad LTV (Gini)  & 0.2612 & \textbf{0.2638} & $+1.0\%$ \\
    \quad CVR (AUROC) & 0.6270 & \textbf{0.6297} & $+0.43\%$ \\
    \addlinespace[2pt]
    \multicolumn{4}{@{}l}{\textit{Offline LLM tokens (in\,/\,out per user; one-time, cached)}} \\
    \quad $\mathbf{Z}_{pre}$ (serving)      & --- & \multicolumn{2}{c}{5{,}790\,/\,451} \\
    \quad $\mathbf{Z}_{post}$ (train only)  & --- & \multicolumn{2}{c}{7{,}084\,/\,453} \\
    \quad $\mathbf{Z}_{fused}$ (train only) & --- & \multicolumn{2}{c}{1{,}906\,/\,906} \\
    \addlinespace[2pt]
    \multicolumn{4}{@{}l}{\textit{Serving latency (ms)}} \\
    \quad P50         & 28.9 & 31.1 & $+7.6\%$ \\
    \quad P99         & 54.5 & 73.1 & $+34\%$ \\
    \bottomrule
    \end{tabular}
\end{table}

All lifts are significant: $\Delta$log-Gini $+0.0141$ (95\% CI $[+0.0027,+0.0259]$) and $\Delta$AUC $+0.0089$ ($[+0.0022,+0.0158]$), under paired bootstrap ($B{=}20{,}000$), DeLong ($p{=}0.0052$), and paired-$t$ on log-MSE ($p{<}10^{-6}$); the UV-CTR guardrail is non-regressing. Profiles are generated offline by a self-hosted GPT-OSS 120B (MXFP4, single 80GB GPU), invoked once on first conversion and cached.

\section{Conclusion}
We presented \modelname{}, a Semantic Reasoning--aware Distillation framework for \textbf{new-user cold-start} prediction in e-commerce. \modelname{} tackles sparsity by (i) using an LLM to densify \textbf{pre-conversion} logs into structured semantic profiles, and (ii) distilling \textbf{post-conversion} privileged signals into teacher targets that are transferable to an online student constrained to pre-conversion inputs. Experiments on a large-scale industrial dataset show consistent gains on both LTV and CVR tasks, and strong \textbf{data efficiency}: using only 9\% of the training data, \modelname{} reaches nearly the same LTV (within 1\%) while improving CVR by 0.8\%. A four-week online A/B test at Keeta confirms these gains in production ($+1.0\%$ LTV, $+0.43\%$ CVR). Finally, the proposed framework is model-agnostic and can be integrated with common industrial backbones.

\bibliographystyle{ACM-Reference-Format}
\bibliography{references}


\begin{thebibliography}{26}


\ifx \showCODEN    \undefined \def \showCODEN     #1{\unskip}     \fi
\ifx \showISBNx    \undefined \def \showISBNx     #1{\unskip}     \fi
\ifx \showISBNxiii \undefined \def \showISBNxiii  #1{\unskip}     \fi
\ifx \showISSN     \undefined \def \showISSN      #1{\unskip}     \fi
\ifx \showLCCN     \undefined \def \showLCCN      #1{\unskip}     \fi
\ifx \shownote     \undefined \def \shownote      #1{#1}          \fi
\ifx \showarticletitle \undefined \def \showarticletitle #1{#1}   \fi
\ifx \showURL      \undefined \def \showURL       {\relax}        \fi
\providecommand\bibfield[2]{#2}
\providecommand\bibinfo[2]{#2}
\providecommand\natexlab[1]{#1}
\providecommand\showeprint[2][]{arXiv:#2}

\bibitem[Bao et~al\mbox{.}(2023)]%
        {bao2023tallrec}
\bibfield{author}{\bibinfo{person}{Ke Bao}, \bibinfo{person}{Jizhi Zhang},
  \bibinfo{person}{Yang Zhang}, \bibinfo{person}{Ruobing Xie}, {and}
  \bibinfo{person}{Xinyu Zhang}.} \bibinfo{year}{2023}\natexlab{}.
\newblock \bibinfo{title}{{TALLRec}: An Effective and Efficient Tuning
  Framework to Align Large Language Model with Recommendation}.
\newblock \bibinfo{howpublished}{arXiv preprint arXiv:2305.12300}.
\newblock


\bibitem[Gao et~al\mbox{.}(2023)]%
        {gao2023chatrec}
\bibfield{author}{\bibinfo{person}{Chunfeng Gao}, \bibinfo{person}{Yongfeng
  Zhang}, \bibinfo{person}{Fuzhen Zhuang}, \bibinfo{person}{Yanchi Liu},
  \bibinfo{person}{Haoyuan Liu}, \bibinfo{person}{Xiangnan He}, {and}
  \bibinfo{person}{Qing Li}.} \bibinfo{year}{2023}\natexlab{}.
\newblock \bibinfo{title}{{Chat-REC}: Towards Interactive and Explainable
  Recommendation via Large Language Models}.
\newblock \bibinfo{howpublished}{arXiv preprint arXiv:2305.14524}.
\newblock


\bibitem[Geng et~al\mbox{.}(2022)]%
        {geng2022p5}
\bibfield{author}{\bibinfo{person}{Shijie Geng}, \bibinfo{person}{Shuchang
  Liu}, \bibinfo{person}{Zuohui Fu}, \bibinfo{person}{Ye Yuan},
  \bibinfo{person}{Zhaochun Ren}, {and} \bibinfo{person}{Gerard de Melo}.}
  \bibinfo{year}{2022}\natexlab{}.
\newblock \bibinfo{title}{Personalized Prompt for Recommendation}.
\newblock \bibinfo{howpublished}{arXiv preprint arXiv:2203.13366}.
\newblock


\bibitem[Guo et~al\mbox{.}(2017)]%
        {guo2017deepfm}
\bibfield{author}{\bibinfo{person}{Huifeng Guo}, \bibinfo{person}{Ruiming
  Tang}, \bibinfo{person}{Yunming Ye}, \bibinfo{person}{Zhenguo Li}, {and}
  \bibinfo{person}{Xiuqiang He}.} \bibinfo{year}{2017}\natexlab{}.
\newblock \showarticletitle{DeepFM: A Factorization-Machine based Neural
  Network for {CTR} Prediction}. In \bibinfo{booktitle}{\emph{Proceedings of
  the 26th International Joint Conference on Artificial Intelligence ({IJCAI}
  2017)}}. \bibinfo{pages}{1725--1731}.
\newblock


\bibitem[Hinton et~al\mbox{.}(2015)]%
        {hinton2015distill}
\bibfield{author}{\bibinfo{person}{Geoffrey Hinton}, \bibinfo{person}{Oriol
  Vinyals}, {and} \bibinfo{person}{Jeff Dean}.}
  \bibinfo{year}{2015}\natexlab{}.
\newblock \bibinfo{title}{Distilling the Knowledge in a Neural Network}.
\newblock \bibinfo{howpublished}{arXiv preprint arXiv:1503.02531}.
\newblock


\bibitem[Ji et~al\mbox{.}(2023)]%
        {ji2023survey}
\bibfield{author}{\bibinfo{person}{Ziwei Ji}, \bibinfo{person}{Nayeon Lee},
  \bibinfo{person}{Rita Frieske}, \bibinfo{person}{Tiezheng Yu},
  \bibinfo{person}{Dan Su}, \bibinfo{person}{Yan Xu}, \bibinfo{person}{Etsuko
  Ishii}, \bibinfo{person}{Ye~Jin Bang}, \bibinfo{person}{Andrea Madotto},
  {and} \bibinfo{person}{Pascale Fung}.} \bibinfo{year}{2023}\natexlab{}.
\newblock \showarticletitle{Survey of Hallucination in Natural Language
  Generation}.
\newblock \bibinfo{journal}{\emph{Comput. Surveys}} \bibinfo{volume}{55},
  \bibinfo{number}{12} (\bibinfo{year}{2023}), \bibinfo{pages}{1--38}.
\newblock
\href{https://doi.org/10.1145/3571730}{doi:\nolinkurl{10.1145/3571730}}


\bibitem[Kang et~al\mbox{.}(2020)]%
        {kang2020de}
\bibfield{author}{\bibinfo{person}{SeongKu Kang}, \bibinfo{person}{Junyoung
  Hwang}, \bibinfo{person}{Wonbin Kweon}, {and} \bibinfo{person}{Hwanjo Yu}.}
  \bibinfo{year}{2020}\natexlab{}.
\newblock \showarticletitle{{DE-RRD}: A Knowledge Distillation Framework for
  Recommender System}. In \bibinfo{booktitle}{\emph{Proceedings of the 29th
  {ACM} International Conference on Information and Knowledge Management
  ({CIKM} '20)}}. \bibinfo{pages}{605--614}.
\newblock
\href{https://doi.org/10.1145/3340531.3412005}{doi:\nolinkurl{10.1145/3340531.3412005}}


\bibitem[Kang and McAuley(2018)]%
        {kang2018sasrec}
\bibfield{author}{\bibinfo{person}{Wang-Cheng Kang} {and}
  \bibinfo{person}{Julian McAuley}.} \bibinfo{year}{2018}\natexlab{}.
\newblock \showarticletitle{Self-Attentive Sequential Recommendation}.
\newblock \bibinfo{journal}{\emph{arXiv preprint arXiv:1808.09781}}
  (\bibinfo{year}{2018}).
\newblock


\bibitem[Kim et~al\mbox{.}(2024)]%
        {kim2025exp3rt}
\bibfield{author}{\bibinfo{person}{Jieyong Kim}, \bibinfo{person}{Hyunseo Kim},
  \bibinfo{person}{Hyunjin Cho}, \bibinfo{person}{SeongKu Kang},
  \bibinfo{person}{Buru Chang}, \bibinfo{person}{Jinyoung Yeo}, {and}
  \bibinfo{person}{Dongha Lee}.} \bibinfo{year}{2024}\natexlab{}.
\newblock \bibinfo{title}{Review-driven Personalized Preference Reasoning with
  Large Language Models for Recommendation}.
\newblock \bibinfo{howpublished}{arXiv preprint arXiv:2408.06276}.
\newblock


\bibitem[Kojima et~al\mbox{.}(2023)]%
        {kojima2023largelanguagemodelszeroshot}
\bibfield{author}{\bibinfo{person}{Takeshi Kojima},
  \bibinfo{person}{Shixiang~Shane Gu}, \bibinfo{person}{Machel Reid},
  \bibinfo{person}{Yutaka Matsuo}, {and} \bibinfo{person}{Yusuke Iwasawa}.}
  \bibinfo{year}{2023}\natexlab{}.
\newblock \bibinfo{title}{Large Language Models are Zero-Shot Reasoners}.
\newblock \bibinfo{howpublished}{arXiv preprint arXiv:2205.11916}.
\newblock
\showeprint[arxiv]{2205.11916}~[cs.CL]


\bibitem[Lopez-Paz et~al\mbox{.}(2016)]%
        {lopezpaz2016unifying}
\bibfield{author}{\bibinfo{person}{David Lopez-Paz},
  \bibinfo{person}{L{\'{e}}on Bottou}, \bibinfo{person}{Bernhard
  Sch{\"{o}}lkopf}, {and} \bibinfo{person}{Vladimir Vapnik}.}
  \bibinfo{year}{2016}\natexlab{}.
\newblock \showarticletitle{Unifying distillation and privileged information}.
  In \bibinfo{booktitle}{\emph{International Conference on Learning
  Representations ({ICLR})}}.
\newblock


\bibitem[Ma et~al\mbox{.}(2018)]%
        {ma2018esmm}
\bibfield{author}{\bibinfo{person}{Jiaqi Ma}, \bibinfo{person}{Zhe Zhao},
  \bibinfo{person}{Xinyang Yi}, \bibinfo{person}{Jilin Chen},
  \bibinfo{person}{Lichan Hong}, {and} \bibinfo{person}{Ed~H. Chi}.}
  \bibinfo{year}{2018}\natexlab{}.
\newblock \bibinfo{title}{Entire Space Multi-Task Model: An Effective Approach
  for Estimating Post-Click Conversion Rate}.
\newblock \bibinfo{howpublished}{arXiv preprint arXiv:1804.07931}.
\newblock


\bibitem[Romero et~al\mbox{.}(2015)]%
        {romero2015fitnets}
\bibfield{author}{\bibinfo{person}{Adriana Romero}, \bibinfo{person}{Nicolas
  Ballas}, \bibinfo{person}{Samira~Ebrahimi Kahou}, \bibinfo{person}{Antoine
  Chassang}, \bibinfo{person}{Carlo Gatta}, {and} \bibinfo{person}{Yoshua
  Bengio}.} \bibinfo{year}{2015}\natexlab{}.
\newblock \showarticletitle{FitNets: Hints for Thin Deep Nets}. In
  \bibinfo{booktitle}{\emph{International Conference on Learning
  Representations ({ICLR})}}.
\newblock


\bibitem[Schein et~al\mbox{.}(2002)]%
        {schein2002coldstart}
\bibfield{author}{\bibinfo{person}{Andrew~I. Schein},
  \bibinfo{person}{Alexandrin Popescul}, \bibinfo{person}{Lyle~H. Ungar}, {and}
  \bibinfo{person}{David~M. Pennock}.} \bibinfo{year}{2002}\natexlab{}.
\newblock \showarticletitle{Methods and Metrics for Cold-Start
  Recommendations}. In \bibinfo{booktitle}{\emph{Proceedings of the 25th Annual
  International {ACM} {SIGIR} Conference on Research and Development in
  Information Retrieval ({SIGIR} '02)}}. \bibinfo{pages}{253--260}.
\newblock
\href{https://doi.org/10.1145/564376.564421}{doi:\nolinkurl{10.1145/564376.564421}}


\bibitem[Sclar et~al\mbox{.}(2024)]%
        {sclar2024quantifying}
\bibfield{author}{\bibinfo{person}{Melanie Sclar}, \bibinfo{person}{Yejin
  Choi}, \bibinfo{person}{Yulia Tsvetkov}, {and} \bibinfo{person}{Alane Suhr}.}
  \bibinfo{year}{2024}\natexlab{}.
\newblock \showarticletitle{Quantifying Language Models' Sensitivity to
  Spurious Features in Prompt Design or: How I learned to start worrying about
  prompt formatting}. In \bibinfo{booktitle}{\emph{International Conference on
  Learning Representations ({ICLR})}}.
\newblock


\bibitem[Song(2024)]%
        {song2024smartspaces}
\bibfield{author}{\bibinfo{person}{Yunpeng Song}.}
  \bibinfo{year}{2024}\natexlab{}.
\newblock \bibinfo{title}{Predicting User Behavior in Smart Spaces with
  LLM-Enhanced Logs and Personalized Prompts (Data Description)}.
\newblock \bibinfo{howpublished}{arXiv preprint arXiv:2412.12653}.
\newblock
\href{https://doi.org/10.48550/arXiv.2412.12653}{doi:\nolinkurl{10.48550/arXiv.2412.12653}}


\bibitem[Tsai et~al\mbox{.}(2024)]%
        {tsai2024llmreason}
\bibfield{author}{\bibinfo{person}{Alicia Tsai}, \bibinfo{person}{Adam Kraft},
  \bibinfo{person}{Long Jin}, \bibinfo{person}{Chenwei Cai},
  \bibinfo{person}{Anahita Hosseini}, \bibinfo{person}{Taibai Xu},
  \bibinfo{person}{Zemin Zhang}, \bibinfo{person}{Lichan Hong},
  \bibinfo{person}{Ed~H. Chi}, {and} \bibinfo{person}{Xinyang Yi}.}
  \bibinfo{year}{2024}\natexlab{}.
\newblock \showarticletitle{Leveraging {LLM} Reasoning Enhances Personalized
  Recommender Systems}. In \bibinfo{booktitle}{\emph{Findings of the
  Association for Computational Linguistics: {ACL} 2024}}.
  \bibinfo{pages}{13176--13188}.
\newblock
\href{https://doi.org/10.18653/v1/2024.findings-acl.780}{doi:\nolinkurl{10.18653/v1/2024.findings-acl.780}}


\bibitem[Vapnik and Vashist(2009)]%
        {vapnik2009lupi}
\bibfield{author}{\bibinfo{person}{Vladimir Vapnik} {and}
  \bibinfo{person}{Akshay Vashist}.} \bibinfo{year}{2009}\natexlab{}.
\newblock \showarticletitle{A New Learning Paradigm: Learning Using Privileged
  Information}.
\newblock \bibinfo{journal}{\emph{Neural Networks}} \bibinfo{volume}{22},
  \bibinfo{number}{5--6} (\bibinfo{year}{2009}), \bibinfo{pages}{544--557}.
\newblock
\href{https://doi.org/10.1016/j.neunet.2009.06.042}{doi:\nolinkurl{10.1016/j.neunet.2009.06.042}}


\bibitem[Volkovs et~al\mbox{.}(2017)]%
        {volkovs2017dropoutnet}
\bibfield{author}{\bibinfo{person}{Maksims Volkovs}, \bibinfo{person}{Guangwei
  Yu}, {and} \bibinfo{person}{Tomi Poutanen}.} \bibinfo{year}{2017}\natexlab{}.
\newblock \showarticletitle{DropoutNet: Addressing Cold Start in Recommender
  Systems}. In \bibinfo{booktitle}{\emph{Proceedings of the 11th {ACM}
  Conference on Recommender Systems ({RecSys} '17)}}. \bibinfo{pages}{88--96}.
\newblock


\bibitem[Wang et~al\mbox{.}(2017)]%
        {wang2017dcn}
\bibfield{author}{\bibinfo{person}{Ruoxi Wang}, \bibinfo{person}{Bin Fu},
  \bibinfo{person}{Gang Fu}, {and} \bibinfo{person}{Mingliang Wang}.}
  \bibinfo{year}{2017}\natexlab{}.
\newblock \showarticletitle{Deep \& Cross Network for Ad Click Predictions}. In
  \bibinfo{booktitle}{\emph{Proceedings of the 1st Workshop on Deep Learning
  for Recommender Systems (DLRS at RecSys 2017)}}.
\newblock


\bibitem[Wu et~al\mbox{.}(2023)]%
        {wu2023llmrec}
\bibfield{author}{\bibinfo{person}{Le Wu}, \bibinfo{person}{Xiangnan He},
  \bibinfo{person}{Yunshan Ma}, \bibinfo{person}{Liwei Chen},
  \bibinfo{person}{Shuqin Li}, {and} \bibinfo{person}{Tat-Seng Chua}.}
  \bibinfo{year}{2023}\natexlab{}.
\newblock \bibinfo{title}{A Survey on Large Language Models for Recommender
  Systems}.
\newblock \bibinfo{howpublished}{arXiv preprint arXiv:2305.19860}.
\newblock


\bibitem[Xu et~al\mbox{.}(2020)]%
        {xu2020pfd}
\bibfield{author}{\bibinfo{person}{Chen Xu}, \bibinfo{person}{Quan Li},
  \bibinfo{person}{Junfeng Ge}, \bibinfo{person}{Jinyang Gao},
  \bibinfo{person}{Xiaoyong Yang}, \bibinfo{person}{Changhua Pei},
  \bibinfo{person}{Fei Sun}, \bibinfo{person}{Jian Wu},
  \bibinfo{person}{Hanxiao Sun}, {and} \bibinfo{person}{Wenwu Ou}.}
  \bibinfo{year}{2020}\natexlab{}.
\newblock \showarticletitle{Privileged Features Distillation at Taobao
  Recommendations}. In \bibinfo{booktitle}{\emph{Proceedings of the 26th {ACM}
  {SIGKDD} Conference on Knowledge Discovery and Data Mining ({KDD} '20)}}.
  \bibinfo{pages}{2590--2598}.
\newblock
\href{https://doi.org/10.1145/3394486.3403309}{doi:\nolinkurl{10.1145/3394486.3403309}}


\bibitem[Yuan et~al\mbox{.}(2025)]%
        {yuan2025hapfd}
\bibfield{author}{\bibinfo{person}{Huining Yuan}, \bibinfo{person}{Wenpeng
  Zhang}, \bibinfo{person}{Zijie Hao}, {and} \bibinfo{person}{Zengde Deng}.}
  \bibinfo{year}{2025}\natexlab{}.
\newblock \showarticletitle{Hardness-aware Privileged Features Distillation
  with Latent Alignment for {CVR} Prediction}. In
  \bibinfo{booktitle}{\emph{Proceedings of the 31st {ACM} {SIGKDD} Conference
  on Knowledge Discovery and Data Mining ({KDD} '25)}}.
  \bibinfo{pages}{5182--5193}.
\newblock
\href{https://doi.org/10.1145/3711896.3737231}{doi:\nolinkurl{10.1145/3711896.3737231}}


\bibitem[Yue et~al\mbox{.}(2025)]%
        {yue2025cot4rec}
\bibfield{author}{\bibinfo{person}{Weiqi Yue}, \bibinfo{person}{Yuyu Yin},
  \bibinfo{person}{Xin Zhang}, \bibinfo{person}{Binbin Shi},
  \bibinfo{person}{Tingting Liang}, {and} \bibinfo{person}{Jian Wan}.}
  \bibinfo{year}{2025}\natexlab{}.
\newblock \showarticletitle{CoT4Rec: Revealing User Preferences Through Chain
  of Thought for Recommender Systems}. In \bibinfo{booktitle}{\emph{Proceedings
  of the {AAAI} Conference on Artificial Intelligence}},
  Vol.~\bibinfo{volume}{39}. \bibinfo{pages}{13142--13151}.
\newblock
\href{https://doi.org/10.1609/aaai.v39i12.33434}{doi:\nolinkurl{10.1609/aaai.v39i12.33434}}


\bibitem[Zhou et~al\mbox{.}(2018)]%
        {zhou2018din}
\bibfield{author}{\bibinfo{person}{Guorui Zhou}, \bibinfo{person}{Chengru
  Song}, \bibinfo{person}{Xiaoqiang Zhu}, \bibinfo{person}{Ying Fan},
  \bibinfo{person}{Han Zhu}, \bibinfo{person}{Xiao Ma},
  \bibinfo{person}{Yanghui Yan}, \bibinfo{person}{Junqi Jin},
  \bibinfo{person}{Han Li}, {and} \bibinfo{person}{Kun Gai}.}
  \bibinfo{year}{2018}\natexlab{}.
\newblock \showarticletitle{Deep Interest Network for Click-Through Rate
  Prediction}. In \bibinfo{booktitle}{\emph{Proceedings of the 24th {ACM}
  {SIGKDD} International Conference on Knowledge Discovery \& Data Mining
  ({KDD} '18)}}. \bibinfo{pages}{1059--1068}.
\newblock
\href{https://doi.org/10.1145/3219819.3219823}{doi:\nolinkurl{10.1145/3219819.3219823}}


\bibitem[Zhou et~al\mbox{.}(2019)]%
        {zhou2019dien}
\bibfield{author}{\bibinfo{person}{Guorui Zhou}, \bibinfo{person}{Xiaoqiang
  Zhu}, \bibinfo{person}{Chenru Song}, \bibinfo{person}{Ying Fan},
  \bibinfo{person}{Han Zhu}, \bibinfo{person}{Xiao Ma},
  \bibinfo{person}{Yanghui Yan}, \bibinfo{person}{Junqi Jin},
  \bibinfo{person}{Han Li}, {and} \bibinfo{person}{Kun Gai}.}
  \bibinfo{year}{2019}\natexlab{}.
\newblock \showarticletitle{Deep Interest Evolution Network for Click-Through
  Rate Prediction}. In \bibinfo{booktitle}{\emph{Proceedings of the {AAAI}
  Conference on Artificial Intelligence}}.
\newblock


\end{thebibliography}

\section*{GenAI Usage Disclosure}

We disclose two forms of generative AI usage in this work. \textbf{(i) As a core methodology component.} \modelname{} relies on a large language model (GPT-OSS 120B) as an essential part of its pipeline: the LLM generates the Densified Semantic Profile from pre-conversion logs, the privileged future-behavior profile from post-conversion logs (training-only), and the Hindsight Distillation Target via the Fusion Compiler. All LLM calls and prompts are described in Section~\ref{sec:pipeline} and the Supplementary Material below. \textbf{(ii) Authoring assistance.} The authors used generative AI tools to refine the writing of this paper (grammar, phrasing, and table formatting). All technical claims, experimental designs, and analyses are the original work of the authors and have been verified for correctness.

\clearpage
\section*{Supplementary Material}

\noindent This appendix provides the supplementary material for the paper:
detailed implementation and training settings, full LLM prompts,
feature-engineering and preprocessing details, and example generated
reasoning profiles. Section references of the form ``Section~X'' in the
main text refer to the numbered sections above; the ``\S\,S$k$'' pointers
refer to the sections below.

\renewcommand{\thesection}{S\arabic{section}}
\setcounter{section}{0}
\section{Implementation and Training Details}
\label{app:implementation}

\subsection{Model Architecture}
\label{app:model_architecture}

We implement \modelname{} using PyTorch with the following configurations:
\begin{itemize}
    \item \textbf{Static Encoder:} A DNN based architecture containing a CrossNet (4 layers) and Deep Residual Blocks (2 blocks $\times$ 2 layers, hidden dim=1024) to capture feature interactions. The output dimension is $d_{stat}=128$.
    \item \textbf{Dynamic Encoder:} A Transformer Encoder with 2 layers, 4 attention heads, and a feed-forward dimension of 256. We employ Attention Pooling to compress the sequence into $d_{dyn}=128$.
    \item \textbf{Semantic-Gated Encoder:} We utilize Qwen-0.6B-Emb to generate initial 1024-dimensional embeddings for each of the $K$ reasoning fields. These are processed by field-specific MLPs (Linear-BN-GELU) and projected to $d_{aug}=512$.
    \item \textbf{Multimodal Fusion:} The three views are concatenated ($128+128+512=768$) and fused via a projection MLP to a shared representation of size 128.
    \item \textbf{Prediction Heads:} Two separate DeepTaskHeads (3-layer MLPs) are used for LTV (regression) and CVR (sigmoid classification).
\end{itemize}

\subsection{LLM Protocols}
\label{app:llm_protocols}

For semantic data generation, we utilize the open-source model \textbf{GPT-OSS 120B} as the inference engine for both student densification and teacher hindsight generation. For the text embedding layer in the Semantic-Gated Encoder, we use the pre-trained \textbf{Qwen-0.6B-Embedding} model frozen during training.

\subsection{Training Configuration}
\label{app:training_config}

We train the model for 50 epochs with a batch size of 64 using the AdamW optimizer. The base learning rate is set to $10^{-3}$ accompanied by a 5-epoch linear warmup and cosine annealing decay. Weight decay is set to $5 \times 10^{-4}$ for regularization.

\section{Prompt Engineering Details}
\label{app:prompts}

\subsection{Student Prompt (Densification)}
The following system prompt is used to generate the Semantic Reasoning profile from the sparse $\mathbf{S}_{pre}$ history.

\begin{tcolorbox}[colback=gray!10, colframe=gray!50, title=System Prompt: Student Profile Generation]
\small
\texttt{You are an expert user behavior analyst. Your goal is to infer latent user traits from sparse interaction logs.}

\texttt{Input Context:} \\
\texttt{- User Events: \{sequence\_logs\}} \\
\texttt{- Static Profile: \{user\_profile\}}

\texttt{Task: Analyze the user along the following dimensions:} \\
\texttt{1. Temporal Resolution: Analyze the latency between discovery and action.} \\
\texttt{2. Transaction Magnitude: Evaluate the relative value tier of interacted items.} \\
\texttt{3. Navigation Intent Specificity: Determine the linearity of the user journey.} \\
\texttt{4. Domain Entropy: Measure the distribution variance of interacted categories.}
\\
...

\texttt{For each dimension, provide a specific \textbf{Reasoning Rationale} based on the evidence. Explain WHY you assigned the score.}
\end{tcolorbox}

\subsection{Teacher Prompt (Hindsight Fusion)}
The following prompt is used to generate the ``Ground Truth'' reasoning by fusing the pre-conversion hypothesis with the post-conversion outcome (as described in the Hindsight-Aware Fusion section of the main paper).

\begin{tcolorbox}[colback=gray!10, colframe=gray!50, title=System Prompt: Hindsight Reasoning Fusion]
\small
\texttt{You are an omniscient analyst with access to the user's future actions. Your goal is to resolve ambiguity in the user's initial behavior.}

\texttt{Input Context:} \\
\texttt{- Initial Behavior (Pre): \{pre\_summary\}} \\
\texttt{- Future Outcome (Post): \{post\_summary\}}

\texttt{Task: Re-evaluate the user's pre-conversion traits in light of the post-conversion outcomes.} \\
\texttt{If the user hesitated in pre-conversion but converted in post-conversion, classify as "Cautious High-Intent."} \\
\texttt{Explicitly explain how the future outcome clarifies the initial ambiguity.}
\end{tcolorbox}

\section{Feature Engineering and Preprocessing Details}
\label{app:features}

This section formalizes the mathematical treatment of static and dynamic features used in our framework, detailing the preprocessing transformations applied to ensure numerical stability and semantic preservation.

\subsection{Static Features}
\label{app:static_features}

Static features $\mathbf{F}_{\text{static}} \in \mathbb{R}^{d_s}$ represent time-invariant or slowly-changing user attributes captured at the first-conversion timestamp $t_0$. We partition these features into categorical and numerical subsets based on their cardinality threshold $\tau = 50$.

\paragraph{Categorical Static Features.}
Let $\mathcal{C} = \{c_1, c_2, \ldots, c_m\}$ denote the set of categorical features where each $c_i$ has cardinality $|\mathcal{V}_{c_i}| < \tau$. The categorical feature set encompasses shop identifiers ($\text{shop\_id}$), brand identifiers ($\text{brand\_id}$), geographic area-of-region codes ($\text{aor\_id}$), shop taxonomies ($\text{shop\_type}$), and user marketing segments ($\text{user\_segment}$).

For each categorical feature $c_i$, we construct a vocabulary $\mathcal{V}_{c_i} = \{v_0, v_1, \ldots, v_{|\mathcal{V}_{c_i}|}\}$ from the training corpus, where $v_0$ is reserved for out-of-vocabulary tokens. We map each feature value to an embedding vector via a learnable transformation $\mathbf{E}_{c_i}: \mathcal{V}_{c_i} \rightarrow \mathbb{R}^{d_e}$, where $d_e = 16$ by default. The unified categorical representation is obtained through concatenation:
\begin{equation}
\mathbf{F}_{\text{cat}} = [\mathbf{E}_{c_1}(v_1) \oplus \mathbf{E}_{c_2}(v_2) \oplus \cdots \oplus \mathbf{E}_{c_m}(v_m)] \in \mathbb{R}^{m \cdot d_e}
\end{equation}
where $\oplus$ denotes vector concatenation.

\paragraph{Numerical Static Features.}
Let $\mathcal{N} = \{n_1, n_2, \ldots, n_k\}$ denote numerical features with cardinality $|\mathcal{V}_{n_j}| \geq \tau$. These include user tenure duration ($\text{user\_tenure\_days}$), pre-conversion session counts ($\text{total\_sessions\_pre}$), pre-conversion event counts ($\text{total\_events\_pre}$), average session duration ($\text{avg\_session\_duration}$), and first-visit-to-conversion temporal span ($\text{first\_visit\_to\_conversion\_days}$).

For each numerical feature $n_j \in \mathcal{N}$, we compute population statistics $\mu_{n_j}$ and $\sigma_{n_j}$ from the training distribution and apply $z$-score normalization:
\begin{equation}
n_j' = \frac{n_j - \mu_{n_j}}{\sigma_{n_j}}
\end{equation}
To ensure gradient stability during training, we apply clipping to bound extreme outliers:
\begin{equation}
n_j'' = \text{clip}(n_j', -5, 5) = \max(-5, \min(5, n_j'))
\end{equation}
The normalized numerical features are concatenated to form $\mathbf{F}_{\text{num}} = [n_1'', n_2'', \ldots, n_k''] \in \mathbb{R}^k$.

\subsection{Dynamic Sequence Features}
\label{app:dynamic_features}

Dynamic features encode the temporal behavioral sequence $\mathbf{S}_{pre} = \{\mathbf{e}_1, \mathbf{e}_2, \ldots, \mathbf{e}_T\}$ (pre-conversion) and $\mathbf{S}_{post}$ (post-conversion, available only during training). Each event $\mathbf{e}_t$ at timestep $t$ is represented by a hybrid feature vector comprising categorical and numerical attributes.

\paragraph{Categorical Sequence Features.}
Each event $\mathbf{e}_t$ is characterized by a tuple of categorical features $(\text{type}_t, \text{shop\_type}_t, \text{shop\_aoi}_t)$ where:
\begin{align}
\text{type}_t &\in \mathcal{V}_{\text{type}} \\
\text{shop\_type}_t &\in \mathcal{V}_{\text{shop}} \\
\text{shop\_aoi}_t &\in \mathcal{V}_{\text{aoi}}
\end{align}
The event type vocabulary $\mathcal{V}_{\text{type}}$ encompasses action primitives such as \texttt{page\_view}, \texttt{item\_click}, \texttt{add\_to\_cart}, \texttt{search}, \texttt{filter\_apply}, \texttt{coupon\_collect}, and \texttt{purchase}. The shop type taxonomy $\mathcal{V}_{\text{shop}}$ includes coarse-grained categories (\texttt{fashion}, \texttt{electronics}, \texttt{home\_kitchen}, etc.), while $\mathcal{V}_{\text{aoi}}$ provides fine-grained shop location or brand cluster identifiers.

Vocabularies are constructed from the union of all data splits to ensure complete coverage. Each categorical feature is mapped to a learned embedding space $\mathbb{R}^{d_{seq}}$ where $d_{seq} = 64$, with index 0 reserved for padding tokens. The categorical event representation is then:
\begin{equation}
\mathbf{e}_t^{\text{cat}} = [\mathbf{E}_{\text{type}}(\text{type}_t) \oplus \mathbf{E}_{\text{shop}}(\text{shop\_type}_t) \oplus \mathbf{E}_{\text{aoi}}(\text{shop\_aoi}_t)] \in \mathbb{R}^{3d_{seq}}
\end{equation}

\paragraph{Numerical Sequence Features.}
Each event $\mathbf{e}_t$ additionally contains numerical features $\mathbf{x}_t = (\text{aov}_t, \text{price}_t, \Delta_t, T_t)$ capturing transaction magnitudes and temporal dynamics. Let $\text{aov}_t$ denote the shop's average order value (price tier proxy), $\text{price}_t$ the product unit price, $\Delta_t$ the inter-event time gap, and $T_t$ the cumulative elapsed time from sequence initiation.

To handle the wide dynamic range of these features, we apply log-compression followed by standardization:
\begin{equation}
x_t' = \log(1 + x_t), \quad x_t'' = \frac{x_t' - \mu_x}{\sigma_x}
\end{equation}
where $\mu_x$ and $\sigma_x$ are computed from the training sequence distribution. Missing values are imputed with zero (the post-standardization mean). The normalized numerical event representation is $\mathbf{e}_t^{\text{num}} = [\text{aov}_t'', \text{price}_t'', \Delta_t'', T_t''] \in \mathbb{R}^4$.

\subsection{Sequence Segmentation and Padding}
\label{app:sequence_processing}

\paragraph{Session Segmentation.}
Raw event streams are partitioned into sessions using a temporal threshold $\tau_{\text{session}} = 300{,}000$ milliseconds (5 minutes). Formally, a session boundary is detected between consecutive events $\mathbf{e}_t$ and $\mathbf{e}_{t+1}$ when:
\begin{equation}
\Delta_{t+1} = \text{timestamp}(\mathbf{e}_{t+1}) - \text{timestamp}(\mathbf{e}_t) > \tau_{\text{session}}
\end{equation}
This segmentation preserves the hierarchical temporal structure of user behavior, enabling the model to distinguish between intra-session engagement patterns and inter-session deliberation periods.

\paragraph{Sequence Length Management.}
To accommodate variable-length sequences within fixed-size batches, we impose a maximum sequence length $L_{\max} = 5{,}600$. Let $\mathbf{S} = \{\mathbf{e}_1, \ldots, \mathbf{e}_T\}$ denote a raw sequence with length $T$. The processed sequence $\mathbf{S}'$ and attention mask $\mathbf{M} \in \{0,1\}^{L_{\max}}$ are constructed as:
\begin{equation}
\mathbf{S}' = \begin{cases}
\{\mathbf{e}_{T-L_{\max}+1}, \ldots, \mathbf{e}_T\} & \text{if } T > L_{\max} \text{ (truncation)} \\
\{\mathbf{e}_1, \ldots, \mathbf{e}_T, \mathbf{0}, \ldots, \mathbf{0}\} & \text{if } T \leq L_{\max} \text{ (padding)}
\end{cases}
\end{equation}
\begin{equation}
\mathbf{M}_i = \begin{cases}
1 & \text{if } i \leq \min(T, L_{\max}) \\
0 & \text{otherwise}
\end{cases}
\end{equation}
The attention mask $\mathbf{M}$ is consumed by Transformer layers to prevent attention computation over padding positions, ensuring that $\text{Attention}(\mathbf{Q}, \mathbf{K}, \mathbf{V})$ assigns zero weight to masked positions.

\subsection{Data Splits and Alignment}
\label{app:data_splits}

\paragraph{Temporal Split Strategy.}
To prevent temporal leakage and ensure realistic evaluation, we partition the dataset $\mathcal{D}$ using a strictly chronological split based on first-conversion timestamps. Let $\mathcal{D}_{\text{train}}$, $\mathcal{D}_{\text{val}}$, and $\mathcal{D}_{\text{test}}$ denote the training, validation, and test sets with cardinalities $|\mathcal{D}_{\text{train}}| = 100{,}000$, $|\mathcal{D}_{\text{val}}| = 10{,}000$, and $|\mathcal{D}_{\text{test}}| = 10{,}000$ respectively, yielding a total of $|\mathcal{D}| = 120{,}000$ stratified users. The temporal ordering ensures:
\begin{equation}
\max_{u \in \mathcal{D}_{\text{train}}} t_0^{(u)} < \min_{u \in \mathcal{D}_{\text{val}}} t_0^{(u)} < \min_{u \in \mathcal{D}_{\text{test}}} t_0^{(u)}
\end{equation}
where $t_0^{(u)}$ denotes the first-conversion timestamp for user $u$. This prevents the model from training on future information during validation or testing.

\paragraph{Feature Alignment.}
For each user $u$, we construct a unified feature tuple $(\mathbf{F}_{\text{static}}^{(u)}, \mathbf{S}_{\text{pre}}^{(u)}, \mathbf{S}_{\text{post}}^{(u)}, \mathbf{R}^{(u)})$ where $\mathbf{R}^{(u)}$ represents the augmented semantic reasoning features. All components are aligned via the user identifier key, with missing components imputed using zero vectors or padding sequences to ensure consistent batch construction across training and inference stages.

\section{Schema Discovery Prompts and Example Reasoning Profiles}
\label{app:examples}

\subsection{Schema Discovery and Consolidation Prompts}
\label{app:discovery_prompts}

We provide the two prompts used in the \emph{discover--curate--audit} schema-construction workflow (described in the main paper). The \emph{Discover} prompt asks the LLM to propose candidate behavioral dimensions from a stratified batch of users; the \emph{Curate} prompt consolidates the deduplicated candidates into a small set of ranked finalists for expert review.

\begin{tcolorbox}[colback=gray!10, colframe=gray!50, title=System Prompt: Dimension Discovery]
\small
\texttt{You are a behavioral analyst. Given a batch of \{n\} users (each with a sparse interaction log and an LTV/CVR outcome label), propose 3--6 \textbf{behavioral dimensions} that best discriminate high- from low-value users.}

\texttt{Requirements per dimension:} \\
\texttt{- Name (concise, interpretable).} \\
\texttt{- Definition (what behavioral signal it captures).} \\
\texttt{- Rationale (why it is predictive of LTV/CVR).}

\texttt{Favor dimensions that are distinct from one another and computable from the available logs.}
\end{tcolorbox}

\begin{tcolorbox}[colback=gray!10, colframe=gray!50, title=System Prompt: Dimension Consolidation]
\small
\texttt{You are given \{m\} candidate behavioral dimensions (deduplicated across batches). Consolidate them into 6--8 \textbf{ranked finalists}.}

\texttt{For each finalist: merge near-duplicates, give a canonical name and definition, and rank by expected predictive relevance and distinctness. Flag any candidate that is redundant or not computable from interaction logs.}
\end{tcolorbox}

\subsection{Example Generated Reasoning Profiles}
\label{app:profile_examples}

We provide illustrative examples of the generated semantic reasoning profiles for one representative validation user, along two dimensions (\textit{Temporal Resolution}, \textit{Transaction Magnitude}). We present the three reasoning profiles produced by our LLM pipeline: (1) \textbf{$\mathbf{Z}_{pre}$} -- student reasoning from pre-conversion logs only (available at serving time), (2) \textbf{$\mathbf{Z}_{post}$} -- reasoning from post-conversion logs (training-time privileged information), and (3) \textbf{$\mathbf{Z}_{fused}$} -- hindsight fusion that teaches the student what signals to trust in $\mathbf{Z}_{pre}$.

\subsection{Example 1: Student Reasoning from Pre-Conversion ($\mathbf{Z}_{pre}$)}

\begin{tcolorbox}[colback=gray!10, colframe=gray!50, title=Dimension 1: Temporal Resolution]
\small
\textbf{Reasoning Rationale ($\mathbf{Z}_{pre}$):} The user generated 35 events over a 46‑minute window, yielding $<$1 event per minute – well below the $>$6 events/minute threshold for a short decision. The activity is spread across 21 sessions, many of which end with app\_quit, indicating a prolonged, fragmented decision process.

\textbf{Label:} Low Resolution (Prolonged Deliberation)
\end{tcolorbox}

\begin{tcolorbox}[colback=gray!10, colframe=gray!50, title=Dimension 2: Transaction Magnitude]
\small
\textbf{Reasoning Rationale ($\mathbf{Z}_{pre}$):} The median price is below the regional average, and 71\% of the price bucket share falls in the lower range. Although the 90th percentile reaches higher values, the overall consumption pattern leans toward lower‑priced items, indicating low transaction magnitude.

\textbf{Label:} Low Magnitude (Budget-Tier)
\end{tcolorbox}

\subsection{Example 2: Teacher Reasoning from Post-Conversion ($\mathbf{Z}_{post}$)}

\begin{tcolorbox}[colback=gray!10, colframe=gray!50, title=Dimension 1: Temporal Resolution]
\small
\textbf{Reasoning Rationale ($\mathbf{Z}_{post}$):} The user generated 110 events over a span of about 7.8 hours ($\approx$468 minutes) across 28 sessions, yielding roughly 0.23 events per minute and only $\sim$3.9 events per session. This low events‑per‑minute ratio and the large number of separate sessions indicate a prolonged decision‑making process rather than a quick, single‑session purchase.

\textbf{Label:} Low Resolution (Prolonged Deliberation)
\end{tcolorbox}

\begin{tcolorbox}[colback=gray!10, colframe=gray!50, title=Dimension 2: Transaction Magnitude]
\small
\textbf{Reasoning Rationale ($\mathbf{Z}_{post}$):} The median price of items the user interacted with is well above the regional average, and the 90th‑percentile price is also substantially higher. Moreover, over 60\% of the price‑bucket share falls into the premium tier, showing a clear preference for higher‑priced products, which signals strong purchasing capacity.

\textbf{Label:} High Magnitude (Premium-Tier)
\end{tcolorbox}

\subsection{Example 3: Hindsight-Aware Fusion ($\mathbf{Z}_{fused}$)}

\begin{tcolorbox}[colback=gray!10, colframe=gray!50, title=Dimension 1: Temporal Resolution]
\small
\textbf{Reasoning Rationale ($\mathbf{Z}_{fused}$):} The low event‑per‑minute rate observed in $\mathbf{S}_{pre}$ ($\approx$0.015) correctly foreshadowed the prolonged deliberation revealed in $\mathbf{S}_{post}$ ($\approx$0.05) and the large number of sessions. Both stages show sparse activity spread over many hours, so the temporal resolution is genuinely slow. \emph{Teaching signal:} The teacher should teach the student to focus on the density of events (events/min) and the count of distinct sessions as primary predictors of Temporal Resolution.

\textbf{Label:} Low Resolution (Prolonged Deliberation)
\end{tcolorbox}

\begin{tcolorbox}[colback=gray!10, colframe=gray!50, title=Dimension 2: Transaction Magnitude]
\small
\textbf{Reasoning Rationale ($\mathbf{Z}_{fused}$):} The high price metrics in $\mathbf{S}_{pre}$ (median and 90th‑percentile well above regional average) were misleading noise caused by a tiny sample of expensive items from a single shop. In $\mathbf{S}_{post}$ the median price drops to well below regional average and the bulk of activity lies in lower price buckets, revealing low transaction magnitude. \emph{Teaching signal:} The teacher should teach the student to give weight to the overall price distribution across all sessions rather than isolated high‑price spikes, and to treat a concentrated high‑price exposure as a weak signal.

\textbf{Label:} Low Magnitude (Budget-Tier)
\end{tcolorbox}

\section{Extended Related Work}
\label{app:related_work}

This section expands the discussion condensed in the main paper's Related Work.

\subsection{Cold-Start Prediction}
Classical collaborative-filtering and ID-based sequence models (DIN~\cite{zhou2018din}, DIEN~\cite{zhou2019dien}, SASRec~\cite{kang2018sasrec}) and feature-interaction models (DeepFM~\cite{guo2017deepfm}, DCN~\cite{wang2017dcn}) achieve strong warm-start accuracy but rely on dense interaction histories; under extreme sparsity their predictions regress toward population priors. Cold-start remedies fall into three families: (i) \emph{robustness} methods such as DropoutNet~\cite{volkovs2017dropoutnet} that randomly mask inputs to reduce over-reliance on dense features; (ii) \emph{meta-learning} approaches that learn fast adaptation from few interactions; and (iii) \emph{self-supervised} sequence objectives that pre-train representations. These mitigate the symptom of sparsity but do not inject external knowledge nor exploit privileged training-time outcomes---the two complementary levers \modelname{} targets.

\subsection{LLMs for Recommendation and Reasoning}
LLMs have been applied to recommendation through prompting and in-context formulations~\cite{geng2022p5,gao2023chatrec}, instruction tuning~\cite{bao2023tallrec}, and reasoning-centric pipelines that elicit natural-language rationales~\cite{yue2025cot4rec,tsai2024llmreason,kim2025exp3rt,wu2023llmrec}. A recurring limitation is that free-form generations are sensitive to prompt phrasing~\cite{sclar2024quantifying} and prone to hallucination~\cite{ji2023survey}, which complicates validation and monitoring in production. \modelname{} constrains generation to a curated behavioral schema, yielding structured, consistent rationales that are encoded by a frozen embedder at serving time so no LLM call is needed online.

\subsection{Privileged Information, Distillation, and Mixtures of Experts}
Learning using privileged information (LUPI)~\cite{vapnik2009lupi,lopezpaz2016unifying} formalizes training-time-only signals, and is commonly realized via knowledge distillation~\cite{hinton2015distill,romero2015fitnets}. In industrial recommendation, privileged-feature distillation (PFD~\cite{xu2020pfd}) and its hardness-aware extension (HAPFD~\cite{yuan2025hapfd}) feed post-conversion features to a teacher, but transfer through a single shared pathway. \modelname{} differs in two ways: (i) it reconciles pre- and post-conversion reasoning into a self-consistent hindsight target $\mathbf{Z}_{fused}$ rather than concatenating contradictory signals, and (ii) it routes transfer through per-user Distillation Experts~\cite{kang2020de}, a mixture-of-experts mechanism that allocates reconstruction capacity to heterogeneous transfer regimes.

\end{document}